\documentclass[10pt,journal,compsoc]{IEEEtran}
\usepackage{times}
\usepackage{epsfig}
\usepackage{graphicx}
\usepackage{amsfonts}
\usepackage{amsmath}
\usepackage{amssymb}
\usepackage{bbm}
\usepackage{multirow}
\usepackage{makecell}
\usepackage{ulem}
\usepackage{booktabs}
\newcommand{\ignore}[1]{}
\usepackage[pagebackref=false,breaklinks=false,colorlinks,bookmarks=false]{hyperref}
\usepackage[table]{xcolor}
\definecolor{cGreen}{RGB}{100,180,100}

\begin{document}
\title{High-Performance Transformer Tracking}
\author{Xin~Chen,
        Bin~Yan,
        Jiawen~Zhu,
        Huchuan~Lu, 
        Xiang~Ruan,
        and~Dong~Wang
\IEEEcompsocitemizethanks{
     \IEEEcompsocthanksitem {Xin Chen, Bin Yan, Jiawen Zhu, Huchuan Lu, and Dong Wang are with the School of Information and Communication Engineering, Dalian University of Technology, Dalian 116024, China. Xin Chen, Huchuan Lu, and Dong Wang are also with the Ningbo Institute, Dalian University of Technology, Ningbo 315016, China.} 
     \IEEEcompsocthanksitem {Xiang Ruan is with the Tiwaki Co., Ltd., Kusatsu, Shiga 525-8577, Japan.}
     \IEEEcompsocthanksitem {Corresponding author: Dong Wang, wdice@dlut.edu.cn}
     }
}
\IEEEtitleabstractindextext{%
\begin{abstract}
%
Correlation has a critical role in the tracking field, especially in recent popular Siamese-based trackers.
The correlation operation is a simple fusion method that considers the similarity between the template and the search region.
However, the correlation operation is a local linear matching process, losing semantic information and easily falling into a local optimum, which may be the bottleneck in designing high-accuracy tracking algorithms.
In this work, to determine whether a better feature fusion method exists than correlation, a novel attention-based feature fusion network, inspired by the transformer, is presented.
This network effectively combines the template and search region features using attention.
Specifically, the proposed method includes an ego-context augment module based on self-attention and a cross-feature
augment module based on cross-attention.
First, we present a transformer tracking (named TransT) method based on the Siamese-like feature
extraction backbone, the designed attention-based fusion mechanism, and the classification and regression head.
Based on the TransT baseline, we further design a segmentation branch to generate an accurate mask.
Finally, we propose a stronger version of TransT by extending TransT with a multi-template scheme and an IoU prediction head, named TransT-M.
Experiments show that our TransT and TransT-M methods achieve promising results on seven popular datasets.
Code and models are available at \href{https://github.com/chenxin-dlut/TransT-M}{https://github.com/chenxin-dlut/TransT-M}.
\end{abstract}

}

\maketitle

\IEEEdisplaynontitleabstractindextext

\IEEEpeerreviewmaketitle

\IEEEraisesectionheading{\section{Introduction}\label{sec:introduction}}

\IEEEPARstart{V}{isual} object tracking is a fundamental task in computer vision that predicts the position and shape of a given target in each video frame.
It has a wide range of applications in robot vision, video surveillance, and unmanned driving.
The main challenges of tracking are large occlusion, severe deformation, and interference from similar objects.
Many efforts have been made in recent years~\cite{DVT-Review,DVT-review-new}; however, designing
a high-accuracy and real-time tracker is still a challenging task.

\begin{figure*}[!h]
\begin{center}
\resizebox{0.75\linewidth}{!}{
\begin{tabular}{c@{}c}
\includegraphics[width=0.45\linewidth]{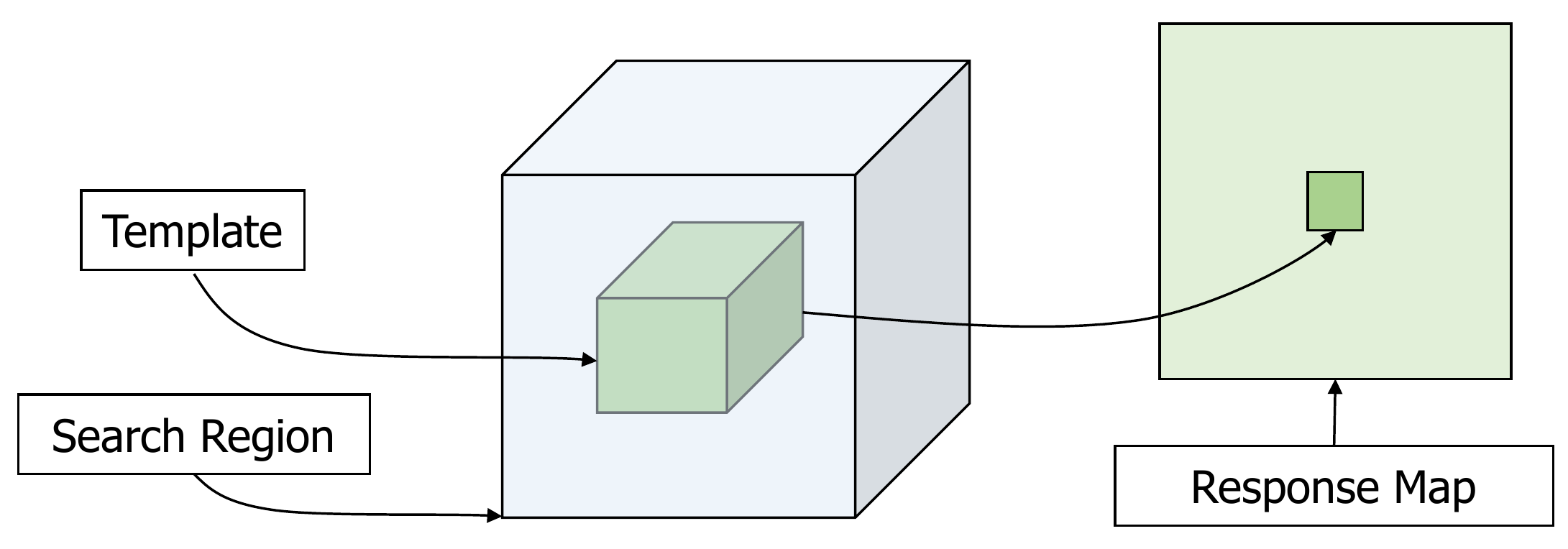} \ &\hspace{20mm}
\includegraphics[width=0.45\linewidth]{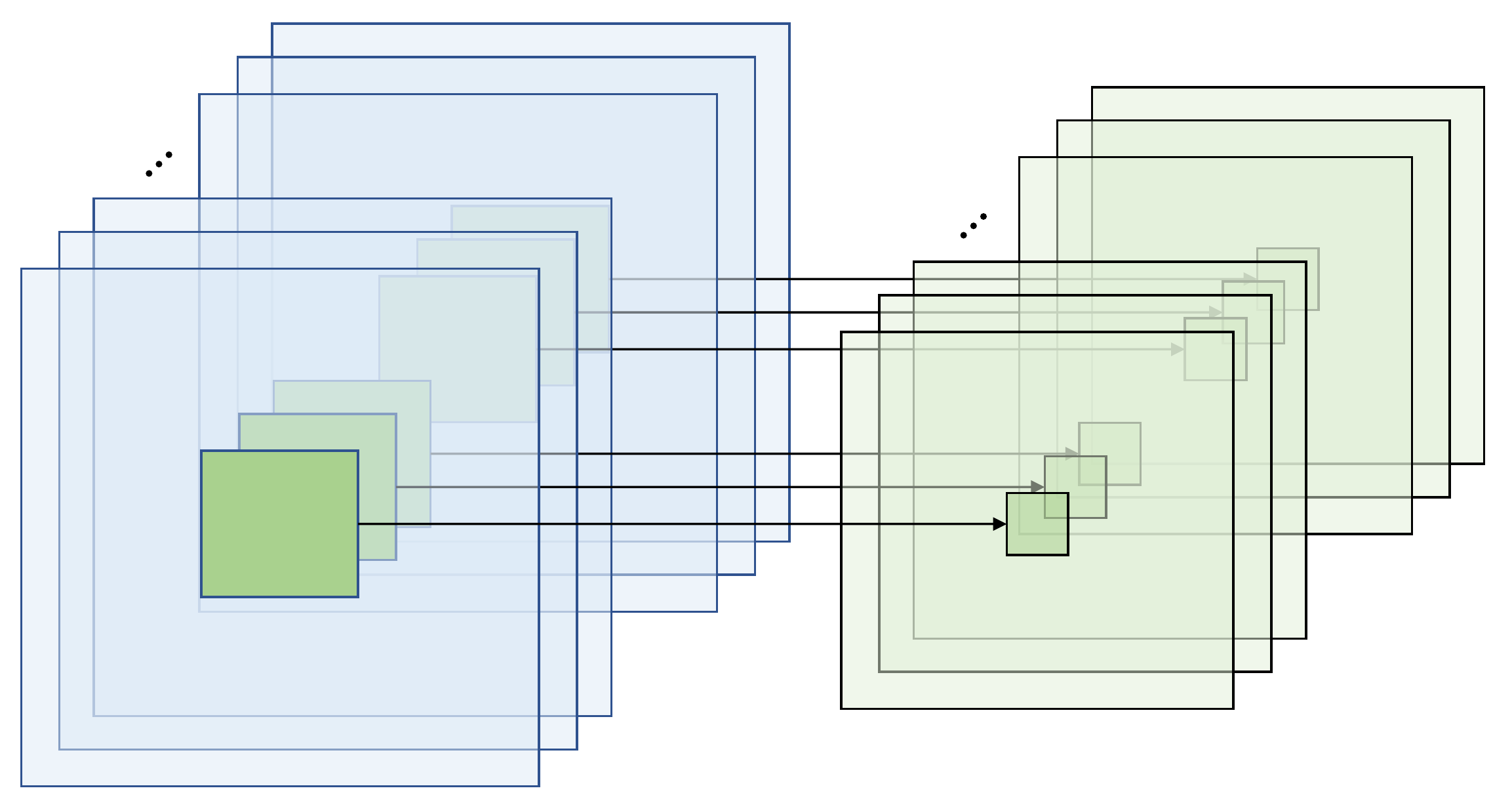} \\
(a) Cross-Correlation Layer &(b) Depthwise Cross-Correlation Layer\\
\end{tabular}}
\end{center}
\vspace{-3mm}
\caption{Cross-correlation layers used in Siamese-based tracking. The left is the original cross-correlation layer and the right is the depthwise cross-correlation layer.}
\label{fig:correlation}
\end{figure*}

For most popular trackers (such as SiamFC~\cite{SiameseFC}, SiamRPN~\cite{SiameseRPN}, and ATOM~\cite{ATOM}), the correlation
plays a critical role in integrating the template or target information into the regions of interest (ROIs).
However, the correlation operation is a linear matching process.
As shown in Figure~\ref{fig:correlation}, correlation calculates the similarity between the template feature and the ROI feature in the form of a sliding window and linear multiplication.
This leads to semantic information loss,
which limits the tracker to capturing the complicated nonlinear interaction between the template and ROIs.
Thus, previous models improve the nonlinear representation ability by introducing fashion structures~\cite{SiamRPNplusplus,SiamFC++,Ocean}, using additional modules~\cite{MLT,DSA,CGACD},
and designing effective online updaters~\cite{DiMP,UpdateNet,PrDiMP}.
Thus, an interesting question is introduced: is there any better feature fusion method than correlation?

\begin{figure}[!t]
\begin{center}
\resizebox{\linewidth}{!}{
\begin{tabular}{p{2cm}<{\centering}@{}p{4.2cm}<{\centering}@{}p{2cm}<{\centering}}
\scriptsize{Self-attention} & \scriptsize{Screenshots} & \scriptsize{Cross-attention}\\
\end{tabular}}
\includegraphics[width=1\linewidth]{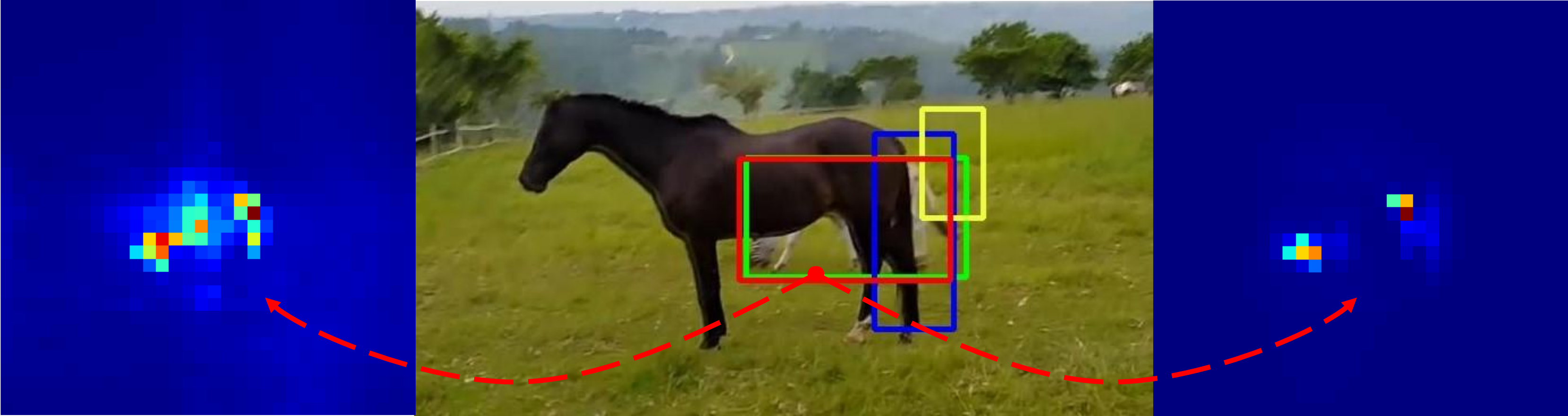}
\end{center}
\vspace{-0.40cm}
\begin{center}
\includegraphics[width=1\linewidth]{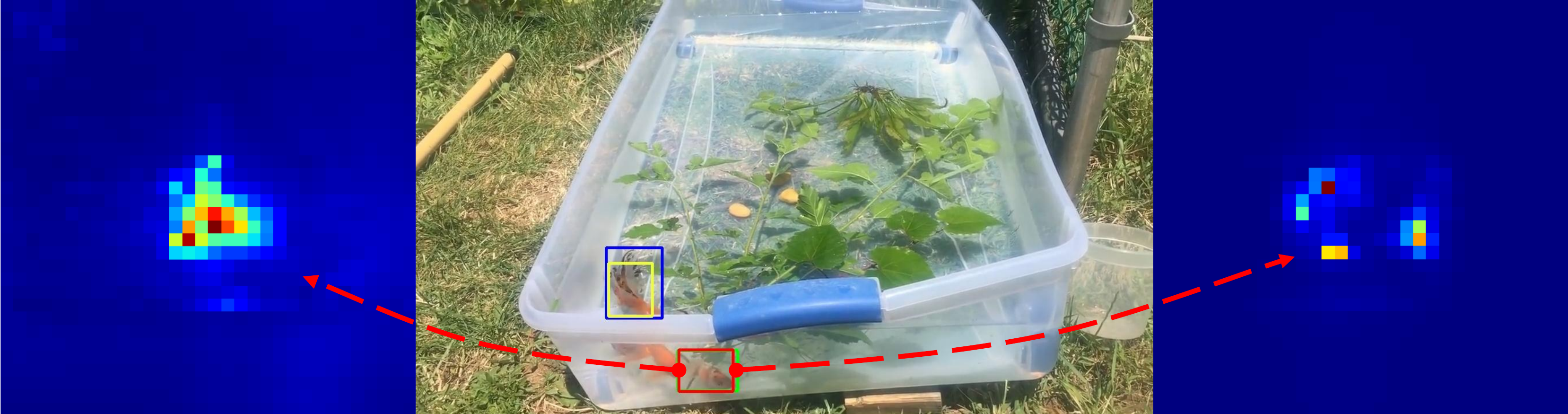}
\end{center}
\vspace{-0.40cm}
\begin{center}
\includegraphics[width=1\linewidth]{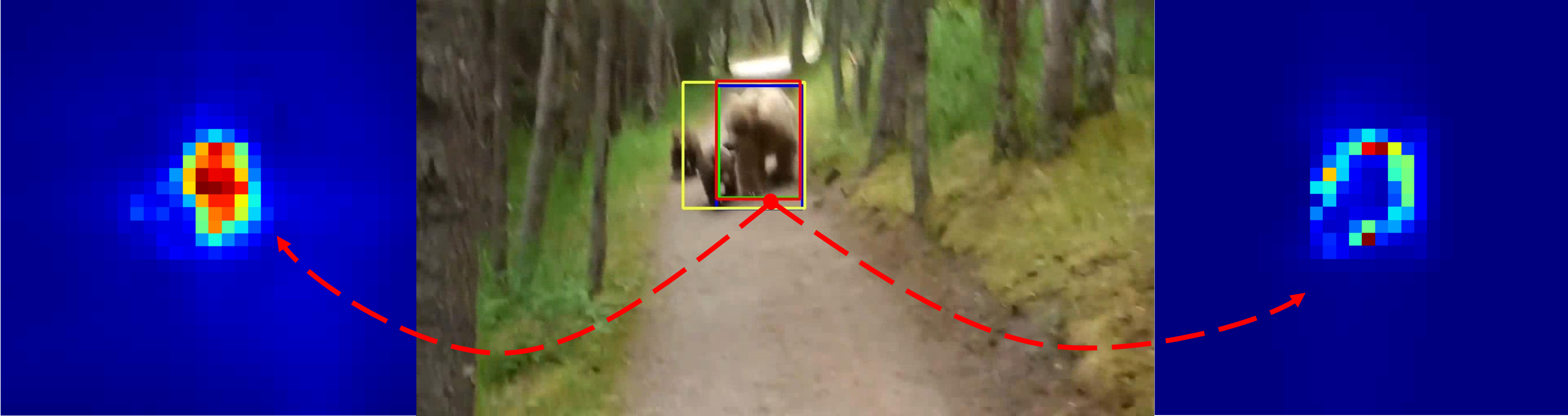}
\end{center}
\vspace{-0.40cm}
\begin{center}
\includegraphics[width=1\linewidth]{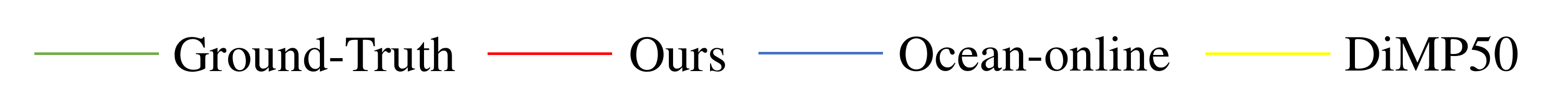}
\end{center}
\vspace{-5mm}
\caption{Tracking results of TransT and two state-of-the-art trackers (DiMP~\cite{DiMP} and Ocean~\cite{Ocean}). Our tracker is more robust
and accurate in handling various challenges, such as occlusion, similar object interference, and motion blur.}
\label{fig:visual}
\end{figure}

Recently, transformers~\cite{2017Attention} have achieved great success in computer vision~\cite{DETR},~\cite{ViT},~\cite{liu2021swin}, benefiting from less inductive bias and global information aggregation ability.
The core of the transformer is the attention mechanism.
The self-attention mechanism integrates global information, enhances effective information, and suppresses redundant information. 
The cross-attention mechanism performs the global nonlinear fusion of input information.
The unrestricted receptive field of attention gives it less inductive bias than convolution and correlation, thus it can easily learn how to perform feature fusion from big data.

In this work, inspired by the core idea of the transformer and attention, we address the bottleneck of the correlation by designing an attention-based feature fusion network and propose a novel transformer tracking
algorithm (named TransT).
The proposed feature fusion network consists of an ego-context augment (ECA) module based on self-attention and a cross-feature augment module (CFA) based on cross-attention.
ECA enhances the single-branch features.
CFA performs feature fusion of two branches, retaining rich semantic information.
ECA and CFA establish dependence between long-distance features and aggregate global information.
This fusion mechanism effectively integrates the template and ROI features, thereby producing more semantic
feature maps than correlation and leading to more accurate tracking results.
Figure~\ref{fig:visual} provides some representative visual results, illustrating that our TransT method produces insightful attention maps regarding the target and performs better than other competing trackers~\cite{Ocean, DiMP}.
%
Given that the previous Siamese trackers~\cite{SiameseFC,SiameseRPN,SiamRPNplusplus,SiamFC++} only use the initial template to compare with the search region, they are not robust to appearance change.
When the target appearance is far from the initial template, the tracker does not work well.
Online methods~\cite{ATOM},~\cite{Ocean},~\cite{DiMP},~\cite{PrDiMP} use historical predictions to update models to handle the appearance change.
Although they have achieved good results, how to select updated samples remains a difficult problem.
In this work, we extend our TransT baseline with a multi-template scheme and an IoU prediction head and propose a stronger tracker called TransT-M.
The multi-template scheme keeps several templates to record the change in the target's appearance.
Specifically, we directly concatenate multiple templates and input them to the template branch.
Benefiting from the characteristic that the transformer can receive variable-length input, our method can flexibly select the number of templates without retraining.
The IoU prediction head predicts the IoU of the predicted bounding box to control the template update.
TransT-M significantly improves the robustness of the tracker.

The combination of tracking and instance segmentation is also a topic of concern. Studies~\cite{SiamMask},~\cite{D3S},~\cite{Alpha-Refine} have predicted the mask and the bounding box. VOT2021~\cite{Kristan_2021_ICCV} uses mask annotation for evaluation. In this work, we also design a segmentation branch for our tracker to generate an accurate mask while predicting the bounding box by a unified framework.

Our main contributions are summarized as follows.

\vspace{-1mm}
\begin{itemize}
    \setlength{\itemsep}{0pt}
    \setlength{\parsep}{0pt}
    \setlength{\parskip}{0pt}
	\item We propose a novel transformer tracking framework (named TransT) consisting of feature extraction, transformer-like fusion,
	and head prediction modules. Transformer-like fusion combines the template and search region features
	solely using attention, without correlation.
	\item We develop our feature fusion network based on an ego-context augment module with self-attention and a cross-feature augment module with cross-attention. Compared with correlation-based feature fusion,
	our attention-based method adaptively focuses on useful information, such as edges and similar targets, and establishes associations between distant features,
	to make the tracker obtain better classification and regression results.

	\item Benefiting from the characteristics of the transformer, we model appearance changes through the multi-template scheme and the IoU prediction head, greatly improving TransT-M for difficult video sequences. We also design a segmentation branch to generate the accurate mask.
    \item Numerous experimental results on many benchmarks show that the proposed tracker performs significantly better than
    the state-of-the-art algorithms, especially on large-scale LaSOT, TrackingNet, GOT-10k, and VOT2021 dataset. In addition, our TransT and TransT-M methods run at approximately 54 and 42 $fps$, respectively, on a GPU, which meets the real-time requirement.
    \item We design many TransT and TransT-M versions using different backbones and feature fusion networks. The fastest version reaches 70 $fps$ with high performance.
\end{itemize}

This study builds upon our conference paper~\cite{TransT} and significantly extends it in various aspects.
First, we design a segmentation branch for TransT to complete bounding box prediction and instance segmentation through a unified framework.
Second, we further extend TransT with the multi-template scheme and the IoU prediction design.
The multi-template scheme records the temporal information and the appearance change of the target.
The IoU prediction head controls the template update.
Third, we provide more variant versions of our method with different numbers of feature fusion layers and backbone networks.
The fastest version reaches 70 $fps$ with high performance.
Furthermore, we incorporate the challenging tracking dataset VOT for more comprehensive evaluations and add more trackers for comparison.

\section{Related Work}\label{sec:related work}

{\noindent \textbf{Visual Object Tracking.}}
In recent years, Siamese-based methods have been more popular in the tracking field~\cite{SINT,SiameseFC,SiameseRPN,SiamMask,SiamRPNplusplus,SiamFC++,Ocean}.
SiamFC~\cite{SiameseFC}, the pioneering work, combined naive feature correlation with a Siamese framework.
After that, SiamRPN~\cite{SiameseRPN} combined the Siamese network with RPN~\cite{FasterRCNN} and conducted feature fusion using
depthwise correlation to obtain more precise tracking results.
Some further improvements have been made, such as adding additional branches~\cite{SiamMask,Alpha-Refine}, using deeper architectures~\cite{SiamRPNplusplus},
and exploiting anchor-free architectures~\cite{SiamFC++,Ocean}.
These mainstream tracking architectures can be divided into two parts: a backbone network to extract image features and a correlation-based network to compute the similarity between the template and the search region.
Some popular online trackers (e.g., ECO~\cite{ECO}, ATOM~\cite{ATOM}, and DiMP~\cite{DiMP}) also heavily rely on the correlation
operation.
However, two issues have been overlooked. First, the correlation-based network does not make full use of the global context; thus, falling into the local optimum is easy. Second, through correlation, the semantic information is lost to some degree, which
may lead to an imprecise prediction regarding the target's boundaries.
Therefore, in TransT~\cite{TransT}, we design a variant transformer structure based on attention to replace the correlation-based network
for conducting feature fusion, thereby significantly improving tracking performance.

Besides our work, TMT~\cite{Wang_2021_Transformer} employs the transformer and combined it with SiameseRPN~\cite{SiameseRPN} and DiMP~\cite{DiMP} as a feature enhancement module rather than replace the correlation.
Stark~\cite{Stark} proposes another transformer tracking framework by concatenating the search region and the template. It also employs the corner prediction head to improve the accuracy of the bounding box prediction and the dynamic template to fuse the temporal information.
In this work, we further expand TransT with the segmentation branch, the multi-template scheme, and the IoU prediction head to improve the tracking performance.
\ignore{
\begin{figure*}[!t]
\begin{center}
\includegraphics[width=1\linewidth]{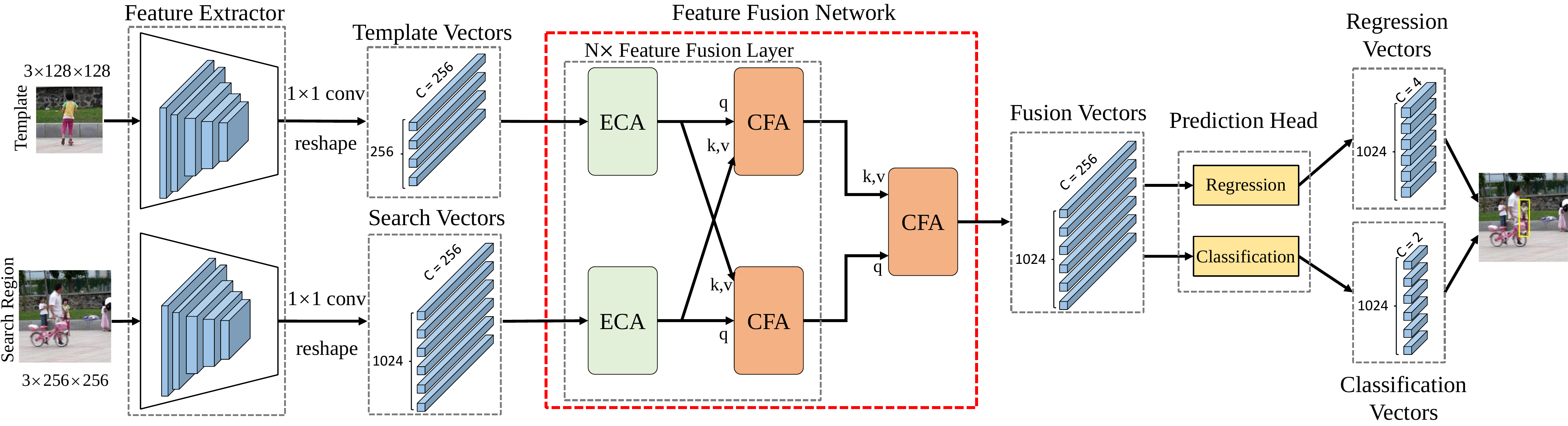}
\end{center}
   \caption{Architecture of our transformer tracking framework. This framework contains three fundamental components:
   feature extraction backbone, feature fusion network, and prediction head. The proposed attention-based feature fusion
   network is naturally applied to the Siamese-based feature extraction backbone.}
   
\label{fig:TransT}
\end{figure*}}

{\noindent \textbf{Transformer and Attention.}} 
Transformer~\cite{2017Attention} was first introduced by Vaswani \emph{et al}. and applied in machine translation.
Briefly, the transformer is an architecture for transforming one sequence into another with the help of attention-based
encoders and decoders.
The attention mechanism looks at an input sequence and decides at each step which other parts of the sequence are important.
Thus, it facilitates capturing the global information from the input sequence. 
The transformer has replaced recurrent neural networks in many sequential tasks (natural language processing~\cite{BERT},
speech processing~\cite{LibriSpeech,ASR}, and computer vision~\cite{ImgTrans}) and gradually extended to handle
nonsequential problems~\cite{ViT,DETR}.
In~\cite{DETR}, Carion \emph{et al}. consider object detection as
a set prediction problem and adopt the encoder-decoder architecture in~\cite{2017Attention} as the detection head.
Experiments on COCO~\cite{COCO} demonstrate that the DETR approach achieves comparable results to an optimized Faster R-CNN
baseline~\cite{FasterRCNN}.
Motivated by the success of DETR and the close relationship between detection and tracking (e.g., RPN~\cite{FasterRCNN} and SiamRPN~\cite{SiameseRPN}), we attempt to introduce the transformer into the tracking field.
Unlike DETR, we do not directly  
follow the encoder-decoder architecture in the original transformer 
because it does not well match the tracking task. 
We adopt the core idea of the transformer and exploit the attention mechanism to design the ego-context augment (ECA) and
cross-feature augment (CFA) modules.
The integration of ECA and CFA focuses on feature fusion between the template and search region rather than extracting
information from only one image in~\cite{DETR}.
This design philosophy is suitable for visual object tracking.

Several efforts have been made to introduce the attention mechanism and transformer in the tracking field.
ACF~\cite{ACF} learns an attention network to do switching among different correlation filters.
MLT~\cite{MLT} adopts channelwise attention to provide the matching network with target-specific information.
These two works merely borrow the concept of attention to conduct model or feature selection.
To improve tracking performance, different attention layers (such as channelwise attention~\cite{RASNet,SA-Siam},
spatial-temporal attention~\cite{CFSTA}, and residual attention~\cite{RASNet}) are utilized to enhance the template
information within the correlation matching framework.
SiamAttn~\cite{DSA} explores self-attention and cross-branch attention to improve the discriminative ability of
target features before applying depthwise cross-correlation.
CGACD~\cite{CGACD} learns attention from the correlation result of the template and search region, and then adopts the learned attention to enhance the search region features for further classification and regression.
These works improved tracking accuracy with the attention mechanism, but they still rely heavily on the
correlation operation in fusing the template and search region features.
Recently, some works~\cite{TransT, Wang_2021_Transformer, Stark} have successfully introduced transformers into tracking and alleviate this dilemma. In this work, we extend our transformer-based tracking method TransT~\cite{TransT}.

\ignore{
Recently, some work has successfully introduced transformers into tracking.
In TransT~\cite{TransT}, we exploited the core idea of the transformer and designed a new attention-based network to fuse the template
and search region features without using any correlation operation.
~\cite{Wang_2021_Transformer} employs a transformer and combines it with SiameseRPN~\cite{SiameseRPN} and DiMP~\cite{DiMP} as a feature enhancement module.
Stark~\cite{Stark} proposed another transformer tracking framework by concatenating the search region and the template.
In this work, we extend and improve TransT. A segmentation branch is designed and integrated into TransT to predict the bounding box and mask through a unified framework. The multitemplate design is employed to record the appearance change and temporal information. The IoU prediction head is used to control the template update.}

\section{Transformer Tracking}
\label{sec:TransT}

\begin{figure*}[!t]
\begin{center}
\includegraphics[width=1\linewidth]{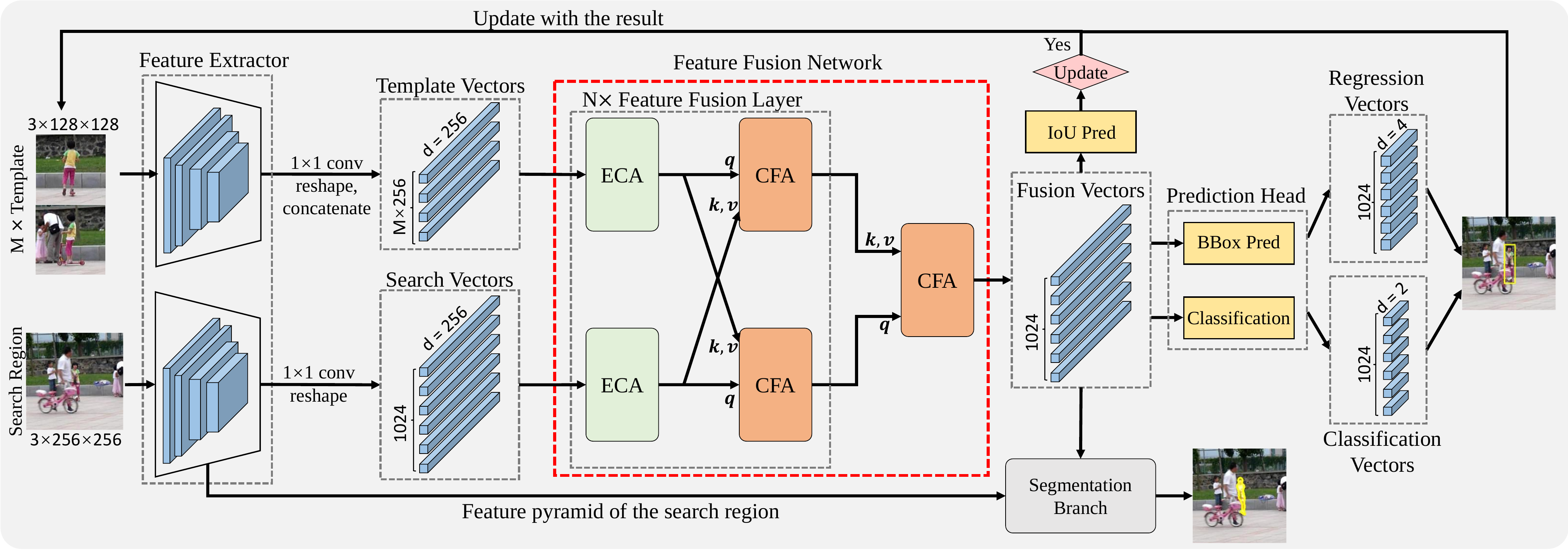}
\end{center}
    \vspace{-5mm}
   \caption{The architecture of our transformer tracking framework. This framework contains three fundamental components: a feature extraction backbone, a feature fusion network, and prediction head networks. Multiple templates are used to record the appearance changes, and an IoU prediction network is used to control the template update.
   Besides, a segmentation branch is used to segment the mask.}
\label{fig:TransT-M}
\end{figure*}

This section presents the proposed transformer tracking methods TransT and TransT-M.
As shown in Figure~\ref{fig:TransT-M}, our TransT consists of three fundamental components:
the backbone network, the feature fusion network, and prediction head networks.
The backbone network separately extracts the features of the template and the search region.
Then, the features are enhanced and fused by the proposed feature fusion network.
Finally, the prediction heads perform the binary classification and bounding box regression on
the enhanced features to generate the tracking results\footnote{The tracking results are also postprocessed
by the window penalty, which will be introduced in Section~\ref{sec-exp}.}.
TransT-M extends TransT with the multi-template scheme and an IoU prediction head. 
Besides, we also design a segmentation branch to generate the accurate mask.
First, we introduce the details of each component of our TransT.
Then, we introduce the two important modules
in the feature fusion network and provide some illustrations and discussions.
Finally, we introduce the details of the segmentation branch and TransT-M.

\subsection{Overall Architecture}
{\noindent \textbf{Feature Extraction. }}
Similar to Siamese-based trackers~\cite{SiameseFC,SiameseRPN}, the proposed TransT method takes a pair of image patches
(i.e., the template image patch $z \in {\mathbb{R}}^{3 \times {H_{z0}} \times {W_{z0}}}$ and the search region 
image patch $x \in {\mathbb{R}}^{3 \times {H_{x0}} \times {W_{x0}}}$) as the inputs of the backbone network. 
The template patch is expanded by twice the side length from the center of the target in the first frame of
a video sequence, which includes the appearance information of the target and its local surrounding scene.
The search region patch is expanded four times the side length from the center coordinate of the target 
in the previous frame, and the search region typically covers the possible moving range of the target. 
The search region and template are reshaped to squares and processed by the backbone. 
We use a modified version of ResNet50~\cite{ResNet} for feature extraction.
Specifically, we remove the last stage of ResNet50 and take the outputs of the fourth stage as final
outputs. We also change the convolution stride of the downsampling unit of the fourth stage from
2 to 1 to obtain a larger feature resolution.
In addition, we modify the $3 \times 3$ convolution in the fourth stage to the dilation convolution with
a stride of 2 to increase the receptive field.
The backbone processes the search region and the template to obtain their feature maps
${f_z} \in {\mathbb{R}}^{C \times {H_z} \times {W_z}}$ and ${f_x} \in {\mathbb{R}}^{C \times {H_x} \times {W_x}}$.
$H_z,W_z = {\frac{H_{z0}}{8}},{\frac{W_{z0}}{8}}$, $H_x,W_x = {\frac{H_{x0}}{8}},{\frac{W_{x0}}{8}}$ and $C = 1024$.

{\noindent \textbf{Feature Fusion Network. }}
We design a feature fusion network to fuse the features $f_z$ and $f_x$. First, a $1 \times 1$ convolution reduces the channel dimension of $f_z$ and $f_x$, obtaining two lower dimension feature maps,  $f_{z0} \in \mathbb{R}^{d \times H_z \times W_z}$ and $f_{x0} \in \mathbb{R}^{d \times H_x \times W_x}$. We employ $d=256$ in our implementation. 
Given that the attention-based feature fusion network takes a set of feature vectors as input, we flatten $f_{z0}$ and $f_{x0}$ in the spatial dimension, obtaining $f_{z1} \in \mathbb{R}^{d \times H_{z}W_{z}}$ and $f_{x1} \in \mathbb{R}^{d \times H_{x}W_{x}}$.
$f_{z1}$ and $f_{x1}$ can be regarded as a set of feature vectors of length $d$.  
As shown in Figure~\ref{fig:TransT-M}, the feature fusion network takes $f_{z1}$ and $f_{x1}$ as the inputs to 
the template branch and the search region branch, respectively.
First, two ego-context augment (ECA) modules focus on the useful
semantic context adaptively by multi-head self-attention to enhance feature representation.
Then, two cross-feature augment (CFA) modules receive the feature maps of their own and the other branch
at the same time and fuse these two feature maps through multi-head cross-attention.
In this way, two ECAs and two CFAs form a fusion layer, as shown in the dotted box in Figure~\ref{fig:TransT-M}.
The fusion layer repeats $N$ times, followed by an additional CFA to fuse the feature map of two branches,
decoding a feature map $f \in \mathbb{R}^{d \times H_{x}W_{x}}$.
We employ $N=4$ by default and provide a $N=2$ version indicated as TransT-N2.
The details of the ECA and CFA modules are introduced in Section~\ref{section:CFAECA}.

{\noindent \textbf{Prediction Head Network.}} 
The prediction head consists of a classification branch and a regression branch, where each branch is a three-layer perceptron with hidden dimension $d$ and a $\rm{ReLU}$ activation function. 
For the feature map $f \in \mathbb{R}^{d \times H_{x}W_{x}}$ generated by the feature fusion network, the head makes predictions on 
each vector to obtain $H_{x}W_{x}$ foreground/background classification results and $H_{x}W_{x}$ normalized coordinates with respect
to the search region size.
Our tracker directly predicts the normalized coordinates instead of adjusting the anchor points or anchor boxes, completely discarding the anchor points or anchor boxes based on prior knowledge, thereby making the tracking framework more concise.

\subsection{Ego-Context Augment and Cross-Feature Augment Modules}
\label{section:CFAECA}
{\noindent \textbf{Multi-head Attention. }}
Attention is the fundamental component in designing our feature fusion network. 
Given queries $\mathbf{Q}$, keys $\mathbf{K}$ and values $\mathbf{V}$, the attention 
function is the scale dot-product attention, defined in Equation (\ref{eq-s-att}).
\begin{equation}
    {\rm{Attention}}(\mathbf{Q},\mathbf{K},\mathbf{V})
    = {\rm{softmax}}(\frac{\mathbf{Q}\mathbf{K}^\top}{\sqrt{d_k}})\mathbf{V}, 
\label{eq-s-att}
\end{equation}
where $d_k$ is the key dimensionality. 
Queries $\mathbf{Q}$ and keys $\mathbf{K}$ are two sets of feature vectors with dimension $d_k$, and values $\mathbf{V}$ are a set of feature vectors with dimension $d_v$. The attention score between $\mathbf{Q}$ and $\mathbf{K}$ is obtained by the scaled dot-product operation, and then the attention map is generated by the $softmax$ operation. The algorithm reweighs $\mathbf{V}$ according to the attention map. In this way, according to the correlation between $\mathbf{Q}$ and $\mathbf{K}$, the attention mechanism adaptively focuses on the useful positions in $\mathbf{V}$.

As described in~\cite{2017Attention}, extending the attention mechanism (\ref{eq-s-att})
into multiple heads enable the mechanism to consider various attention distributions and make the model
focus on different information aspects. The mechanism of multi-head attention
is defined in Equation (\ref{eq-m-att}). We refer the reader to the literature~\cite{2017Attention}
for more detailed descriptions.
\begin{equation}
\label{eq-m-att}
{\rm{MultiHead}}(\mathbf{Q},\mathbf{K},\mathbf{V}) = {\rm{Concat}}({\mathbf{H}_1},...,{\mathbf{H}_{{n_h}}}){\mathbf{W}^O}, 
\end{equation}
\begin{equation}
{\mathbf{H}_i} = {\rm{Attention}}(\mathbf{Q}\mathbf{W}_i^Q,\mathbf{K}\mathbf{W}_i^K,
\mathbf{V}\mathbf{W}_i^V),
\end{equation}
where $\mathbf{W}_i^Q \in \mathbb{R}^{d_m \times d_k}$, $\mathbf{W}_i^K \in \mathbb{R}^{d_m \times d_k}$, $\mathbf{W}_i^V \in \mathbb{R}^{d_m \times d_v}$, and $\mathbf{W}^O \in \mathbb{R}^{n_hd_v \times d_m}$ are parameter matrices. 
In the present work, we employ $n_h=8$, $d_m=256$, and 
${{d_k} = {d_v} = {{{d_m}} \mathord{\left/
 {\vphantom {{{d_m}} {{n_h}}}} \right.
 \kern-\nulldelimiterspace} {{n_h}}} = 32}$
as default values.

\begin{figure}[t]
\begin{center}
\includegraphics[width=0.9\linewidth]{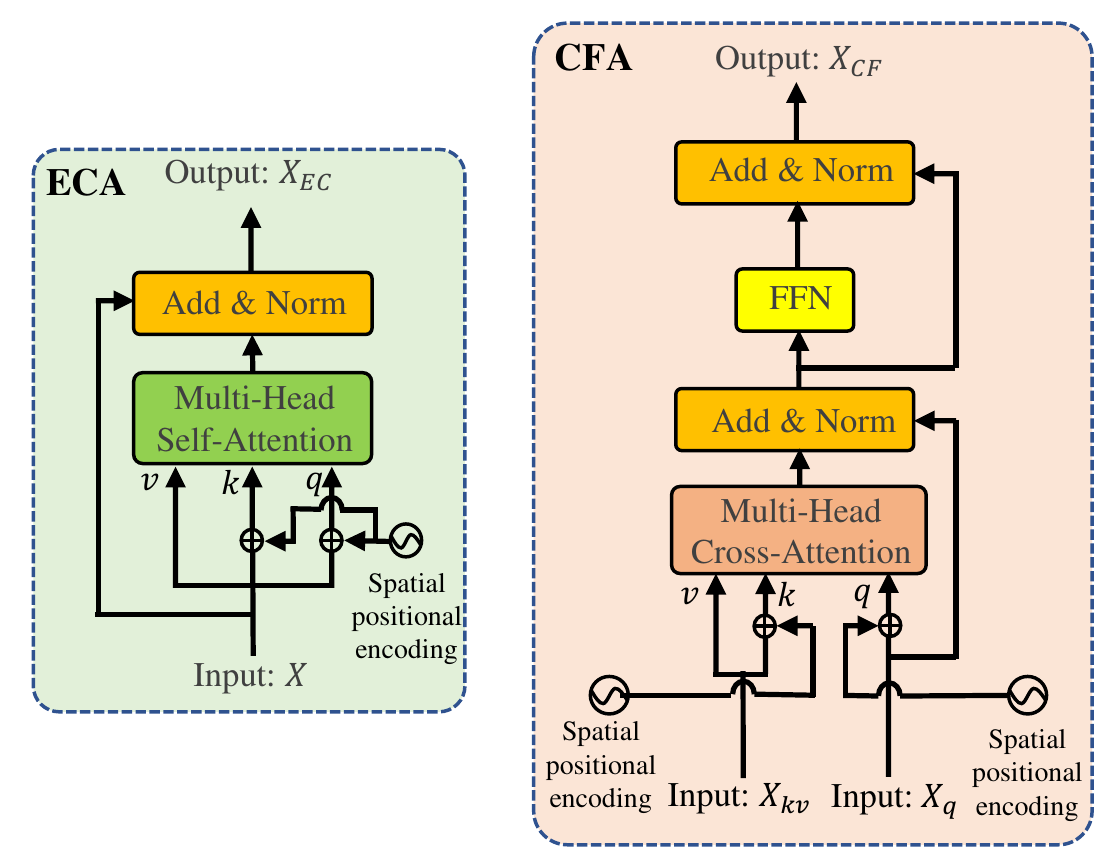}
\vspace{-5mm}
\end{center}
   \caption{Left: ECA module. Right: CFA module. The ECA module is based on multi-head self-attention in a residual form. The CFA module is based on multi-head cross-attention and FFN in a residual form. The input $X_q$ receives the feature from the branch where CFA is located, and $X_{kv}$ receives the feature from the other branch. Spatial positional encodings are used to encode position information. ECA enhances the contextual information of the inputs, and CFA adaptively fuses the features from two branches.}
\label{fig:ECACFA}
\end{figure}

{\noindent \textbf{Ego-Context Augment (ECA). }}
The structure of ECA is shown on the left side of Figure~\ref{fig:ECACFA}.
ECA adaptively integrates the information from different positions of the feature map by using multi-head self-attention in the residual form.
As shown in Equation (\ref{eq-s-att}), the attention mechanism has no ability to distinguish 
the position information of the input feature sequence. 
Thus, we introduce a spatial positional encoding process to the input $\mathbf{X} \in \mathbb{R}^{d \times N_x}$. 
Following \cite{DETR}, we use a sine function to generate spatial positional encoding. 
Finally, the mechanism of ECA can be summarized as 
\begin{equation}
\begin{split}
\label{equation:Multi-Head Attention}
{{\bf{X}}_{EC}} = {\bf{X}} + {\rm MultiHead}({\bf{X}} + {{\bf{P}}_x},{\bf{X}} + {{\bf{P}}_x},{\bf{X}})
\end{split}, 
\end{equation}
where $\mathbf{P}_x \in \mathbb{R}^{d \times N_x}$ is the spatial positional encoding and $\mathbf{X}_{EC} \in
\mathbb{R}^{d \times N_x}$ is the output of ECA.

{\noindent \textbf{Cross-Feature Augment (CFA). }}
The structure of CFA is shown in the right of Figure~\ref{fig:ECACFA} . 
CFA fuses the feature vectors from two inputs by using multihead cross-attention in the form of residuals.
Similar to ECA, spatial positional encoding is also used in CFA. 
Furthermore, an FFN module is used to enhance the fitting ability of the model, which is a fully connected
feed-forward network that consists of two linear transformations with a $\rm{ReLU}$ in between. That is,
\begin{equation}
{\rm FFN}\left( {\bf{x}} \right) = \max \left( {{\bf{0}},{\bf{x}}{{\bf{W}}_1} + 
{{\bf{b}}_1}} \right){{\bf{W}}_2} + {{\bf{b}}_2},
\end{equation}
where $\mathbf{W}$ and $\mathbf{b}$ represent weight matrices and basis vectors, respectively.
The subscripts denote different layers.

Thus, the mechanism of CFA can be summarized as follows: 
\begin{equation}
\begin{array}{c}
{{\bf{X}}_{CF}} = {\widetilde{\bf{X}}_{CF}} + {\rm{FFN}}\left( {{{\widetilde{\bf{X}}}_{CF}}} \right), \\
{\widetilde{\bf{X}}_{CF}} = {{\bf{X}}_q} + {\rm{MultiHead}}\left( {{{\bf{X}}_q} + {{\bf{P}}_q},{{\bf{X}}_{kv}} + {{\bf{P}}_{kv}},{{\bf{X}}_{kv}}} \right), 
\end{array}
\label{eq-cfa}
\end{equation}
where $\mathbf{X}_q \in \mathbb{R}^{d \times N_q}$ is the input of the branch
that the module is applied, $\mathbf{P}_q \in \mathbb{R}^{d \times N_q}$ is the spatial
positional encoding corresponding to $\mathbf{X}_q$,
$\mathbf{X}_{kv} \in \mathbb{R}^{d \times N_{kv}}$ is the input from another branch, $\mathbf{P}_{kv} \in \mathbb{R}^{d \times N_{kv}}$ is the spatial encoding for the coordinate of $\mathbf{X}_{kv}$, and $\mathbf{X}_{CF} \in \mathbb{R}^{d \times N_q}$ is the output of CFA. 
According to Equation (\ref{eq-cfa}), CFA calculates the attention map according to multiple scaled products between $\mathbf{X}_{kv}$ and $\mathbf{X}_q$ and then reweighs $\mathbf{X}_{kv}$ according to the attention map, 
and adds it to $\mathbf{X}_q$ to enhance the representation ability of the feature map.

\subsection{Discussion}
\label{disscussions}

\begin{figure}[t]
\begin{center}
\resizebox{1\linewidth}{!}{
\includegraphics[width=1\linewidth]{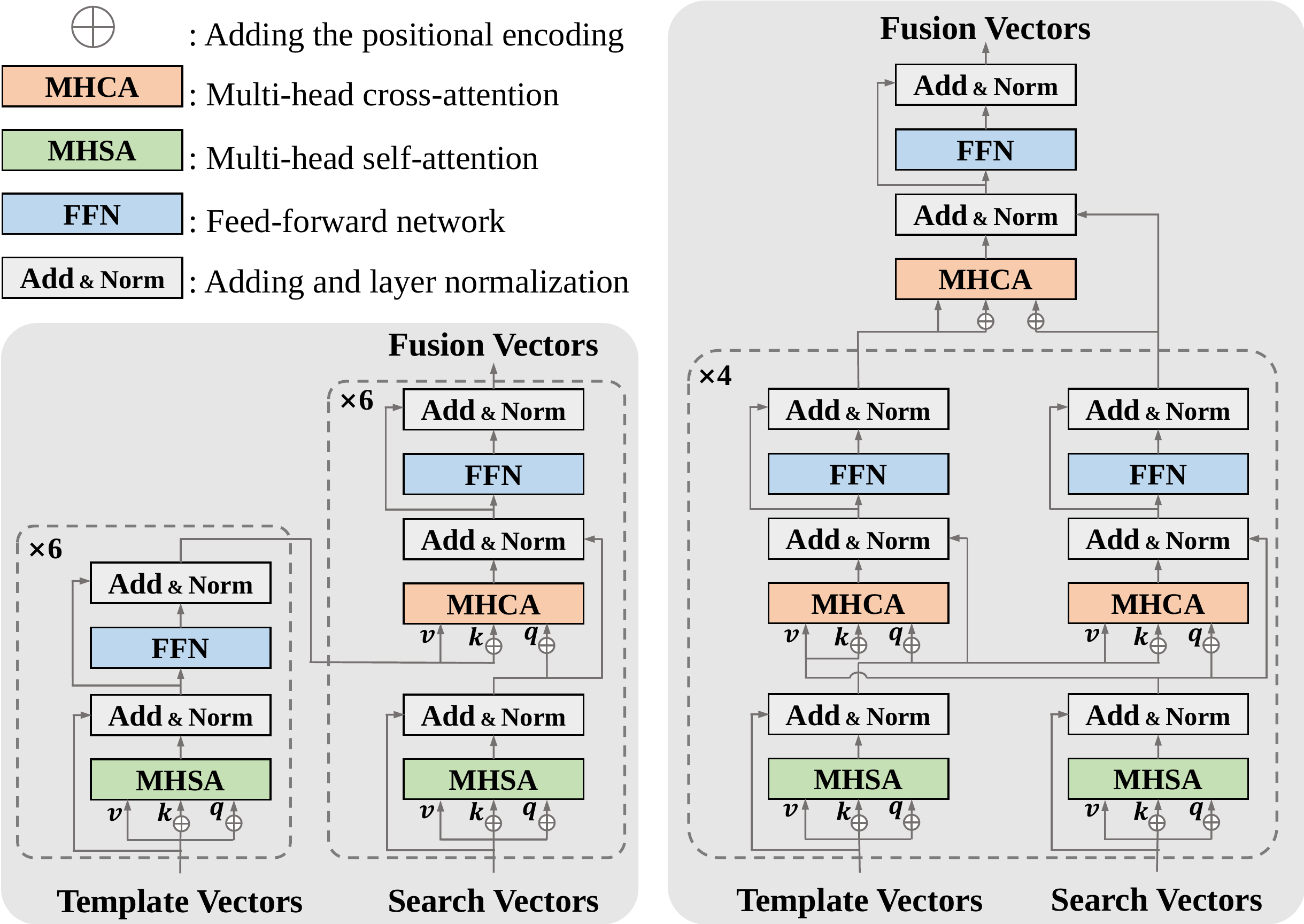}}
\end{center}
\vspace{-3mm}
\caption{Detailed architecture of our feature fusion network and the original transformer. Left: original transformer architecture used in TransT(ori). Right: feature fusion network of TransT.
The parameters of TransT and TransT(ori) are 23.0 and 26.4 M, respectively.}
\label{fig:TransTandOriTransT}
\end{figure}

{\noindent \textbf{Differences from the original transformer. }}
The right of Figure~\ref{fig:TransTandOriTransT} shows the detailed architecture of the feature fusion network of our TransT,
which is a modified version of the transformer.
We also design a tracker employing the original transformer named TransT(ori).
The left of Figure~\ref{fig:TransTandOriTransT} shows the detailed architecture of the original transformer that we employ in TransT(ori).
In TransT(ori), the transformer takes template vectors as the encoder's input and adopts search vectors as the decoder's input rather than the opposite because we found that the performance is better in this way. We provide more discussion about this issue in Section~\ref{subsec-as}.
However, the original transformer is unsuitable for the tracking framework.
First, the transformer is asymmetric, and the processing of the two inputs is inconsistent.
Cross-attention only reweighs the template vectors but not the search vectors, which is unreasonable.
Second, self-attention is used more than cross-attention, which does not match the tracking task well because the interaction between the template and search region is much more important in tracking.
For these reasons, our method draws on the core idea of the transformer, i.e., employing the attention mechanism, but we do not directly adopt the structure of the transformer in DETR~\cite{DETR}.
Instead, we design a new structure to make it more suitable for tracking.
First, our feature fusion network is a symmetrical structure. Thus, it can be naturally integrated into the Siamese framework.
Second, the cross-attention operation in our method plays a more important role than in the original transformer because the tracking task focuses more on fusing the template and search region features.
Although many differences are noted, we still consider our feature fusion network as a variant transformer because it employs the basic structures in the transformer, such as the residual module of the feed-forward network (FFN) and attention. Our method draws on the core idea of the transformer.
Experimental comparisons of the trackers with our method and the original transformer and more discussions
are provided in Section~\ref{subsec-as}.

\begin{figure}[t]
\begin{center}
\includegraphics[width=1\linewidth]{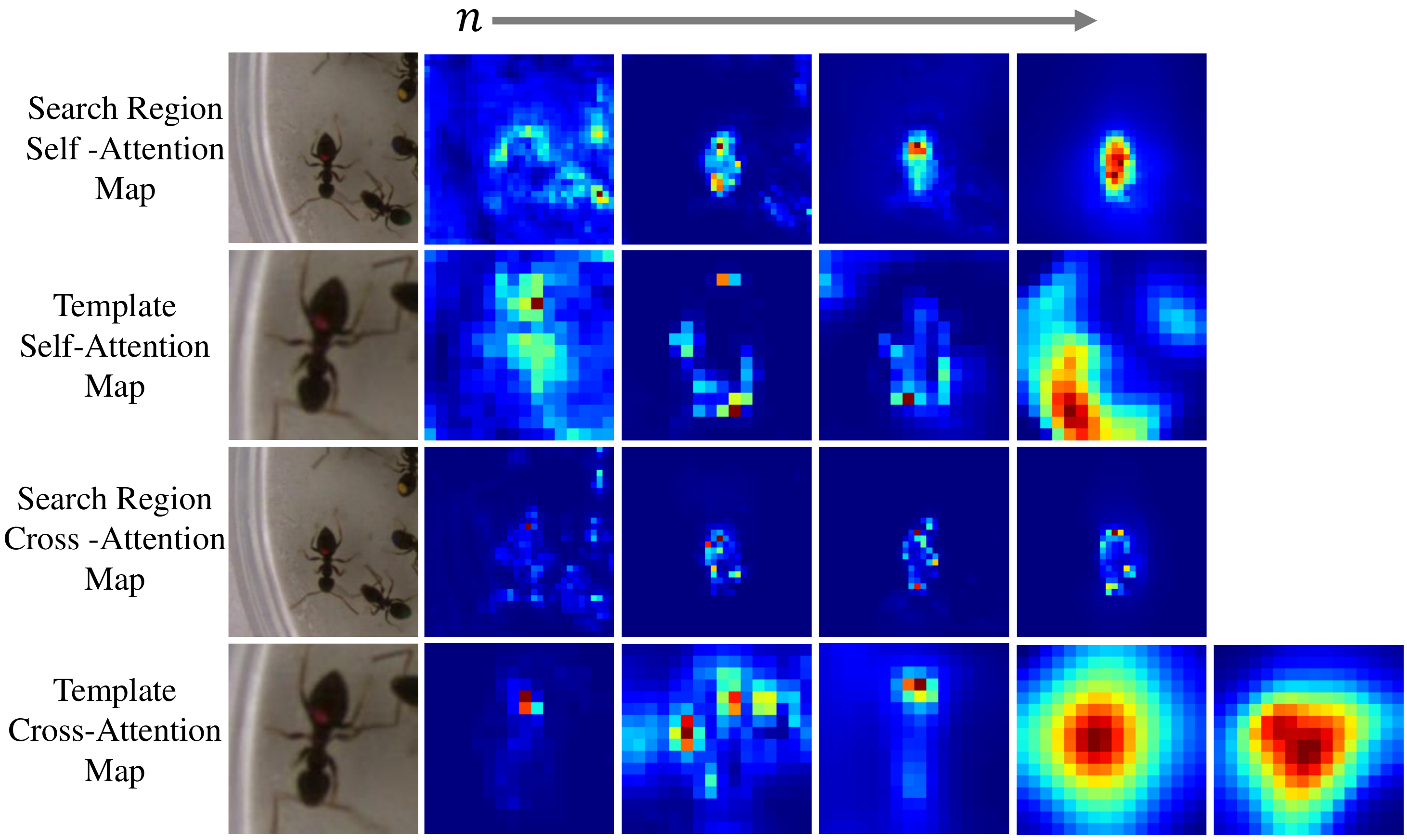}
\end{center}
\vspace{-5mm}
   \caption{Visualization of the attention maps for a representative pair. From left to right, the feature fusion 
   layer goes deeper. From top to bottom, they are self-attention maps in the search region, self-attention maps 
   in the template, cross-attention maps in the search region, and cross-attention maps in the template.}
\label{fig:Attentionmap}
\end{figure}

{\noindent \textbf{What does attention want to see?}} 
To explore how the attention module works in our framework, we visualize the attention maps of all attention modules in a representative tracking clip, as shown in Figure~\ref{fig:Attentionmap}, to see what the attention wants to see. 
Each feature vector has an attention map.
For the attention map from the search region feature vectors, we select the vector with the highest confidence score.
For the attention map from the template feature vectors, we select the vector corresponding to the middle position of the template patch.
We use the number $n$ ($1{\leq}n{\leq}4$) to represent the current number of the fusion layer.
We use four layers in total, and the fusion layer goes deeper from left to right.
The last single attention map is obtained from the last cross-attention, which is used for decoding.

The first line shows self-attention maps of the search region. 
When $n=1$, the template gives no information, and the attention module attempts to see all objects that are different from the environment.
The same thing happens in the second line, that is, the self-attention map of the template.
Attention focuses more on key information, such as the red dot on the tail of the ant.
The third and fourth lines are cross-attention maps applied to the search region and template, respectively.
At this point, attention modules receive features from the template and search region.
To locate the target under the interference of similar targets, attention modules tend to focus on the important information, that is, the colored points on the tail of ants.
When $n=2$, at this point, the inputs of every attention module have fused the search region and template information. 
The focus of the search region self-attention map on similar distractors is reduced, and the model appears to have recognized the target. 
The cross-attention map to the search region seems quite sure of its estimation. 
For the template, attention modules begin to focus on boundary information. 

As the fusion layers become deeper, the search region self-attention map tends to strengthen the location of the target, whereas the cross-attention map to the search region focuses on the boundary of the identified target. 
In this way, the template feature becomes an information bank that contains a large amount of the target's boundary information, and the search region feature still keeps its spatial information. 
The last few attention maps for the template no longer follow the initial spatial position but show a puzzling distribution.
The underlying reason is that after the target is identified, the features of the template branch no longer need to keep the information of the template but store much of the target's boundary information, becoming a feature library serving for regression. Through visualizing the attention maps, one can see that the attention modules automatically look for global useful information, thereby making the tracker achieve good results.

\begin{figure*}[!t]
\begin{center}
\includegraphics[width=0.8\linewidth]{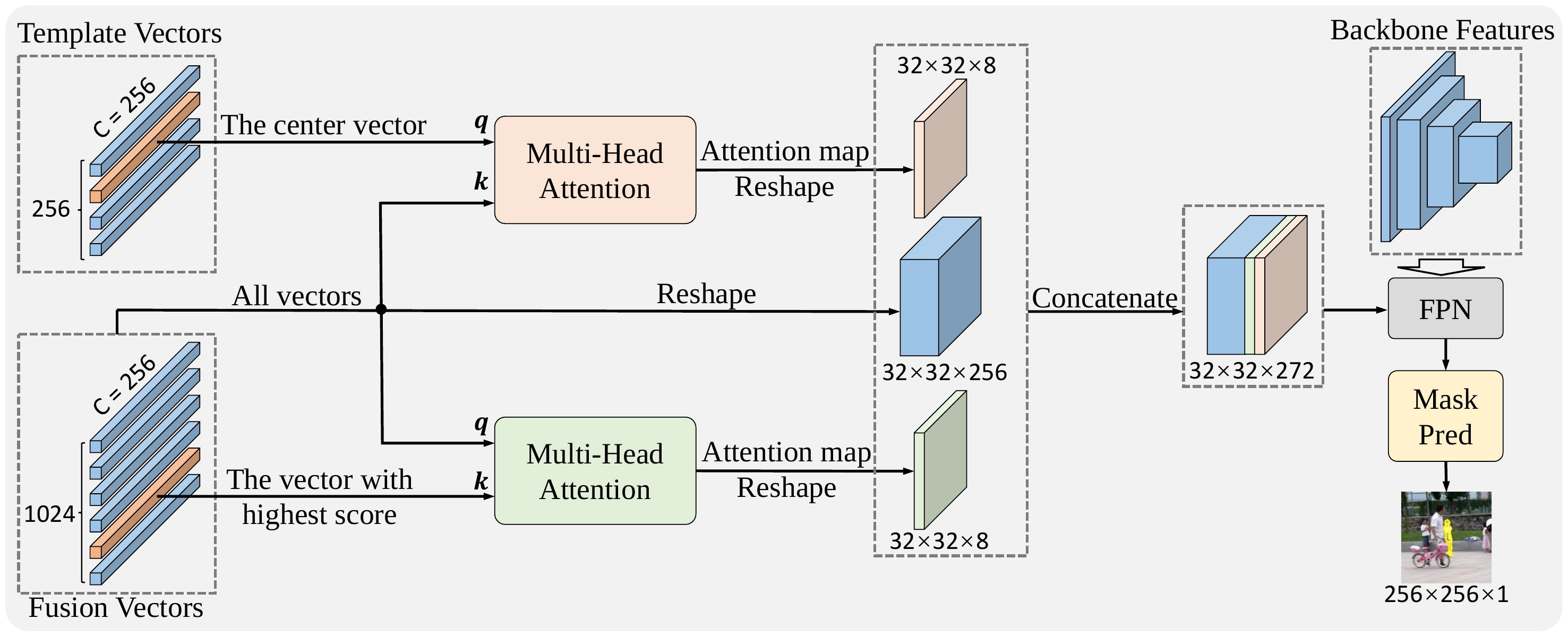}
\vspace{-5mm}
\end{center}
   \caption{The architecture of the segmentation branch. We concatenate the fusion feature with two attention maps, fuse the concatenated feature with the low-level backbone features using the FPN network, and use the mask prediction head to predict the mask. The mask prediction head is a $3\times3$ convolution followed by a bilinear interpolation.}
\label{fig:Seghead}
\end{figure*}

\ignore{
\begin{table*}[!t]
\caption{Comparison with state-of-the-art on VOT2021. The best two results 
are shown in \textbf{\textcolor{red}{red}} and \textbf{\textcolor{blue}{blue}} fonts.}
\label{tab-sota-vot}
\vspace{-6mm}
\begin{center}
\resizebox{\linewidth}{!}{
\begin{tabular}{ccc ccc ccc ccc cc}
\hline
&TransT-M &TransT-SEG &RPT &STARK &OceanPlus &AlphaRef~\cite{Alpha-Refine} &FastOcean\cite{Ocean} &D3S~\cite{D3S} &Ocean~\cite{Ocean} &SiamMargin &SiamEM &STM &SiamMask\\
\hline
EAO    &\textcolor{red}{\textbf{55.2}} &49.9 &\textcolor{blue}{\textbf{53.0}} &50.5 &49.1 &48.2 &46.1 &43.9 &43.0 &35.6 &32.1 &31.0 &30.8\\
Accuracy    &73.9 &\textcolor{blue}{\textbf{75.0}} &70.0 &\textcolor{red}{\textbf{75.9}} &68.5 &75.4 &69.3 &69.9 &69.3 &69.8 &62.4 &52.0 &75.1\\
Robustness &\textcolor{red}{\textbf{87.0}} &80.9 &\textcolor{blue}{\textbf{86.9}} &81.7 &84.2 &77.7 &80.3 &76.9 &75.4 &64.0 &64.8 &74.3 &57.4\\
\hline
\end{tabular}
}
\end{center}
\vspace{-7mm}
\end{table*}
}

\subsection{Segmentation Branch}
\label{sec-seg}
Figure~\ref{fig:Seghead} shows the detailed architecture of the segmentation branch. 
The template vectors are the output feature vectors of the last feature fusion layer's template branch in TransT. 
The fusion vectors are the output feature vectors of the feature fusion network in TransT. 
For the template vectors, we obtain the feature vector corresponding to the middle position of the template patch (indicated as the center vector in Figure~\ref{fig:Seghead}) and employ an additional multi-head attention to calculate an attention map from it to the fusion vectors.
We employ eight-head attention to obtain an attention map with dimensions of $32 \times 32 \times 8$.
For the fusion vectors, we obtain the vector with the highest confidence score predicted by the classification head and employ another additional multi-head attention to calculate an attention map from it to the fusion vectors.
Then, we obtain another attention map with the shape of $32 \times 32 \times 8$.
After that, we reshape the fusion vectors back in the spatial dimension to $32 \times 32 \times 256$ and concatenate it with the two attention maps.
We employ FPN~\cite{lin2017feature} to fuse the concatenated feature with the low-level feature pyramid of the search region (from stages 1 to 4 of the backbone network).
The shape of the output feature of the FPN is $128 \times 128 \times 8$.
Finally, we use a mask prediction head to predict the segmentation mask with the shape of $256 \times 256 \times 1$.
The mask prediction head is a $3\times3$ convolutional layer followed by a bilinear interpolation.

\subsection{TransT-M}
\label{sec-transt-m}
As shown in Figure~\ref{fig:TransT-M}, we extend TransT with a multi-template scheme and an IoU prediction head to produce a stronger tracker named TransT-M. The multi-template scheme records different appearances of the target, making the tracker more robust to the appearance change. The IoU prediction head controls the update of the template by comparing the predicted IoU with the update threshold.

{\noindent \textbf{Multi-template Scheme.}}
In the tracking process, sometimes the appearance of the target changes greatly.
Finding the target only by relying on the first frame template, especially in the case of distractors, is difficult for the tracker.
Thus, we use multiple templates to record the target's appearance changes.
Specifically, we record $M$ templates in online tracking.
In the $M$ templates, we always maintain the initial template as it is precise and reliable.
For the rest of the $M-1$ templates, we update them with the template collected during tracking; these templates are not as precise as the initial templates.
We use the backbone to extract the features from these templates.
Then, after reshaping them to 2D features, we concatenate them in the spatial dimension and input the concatenated feature to the template branch of the feature fusion network.
We employ $M = 2$ by default.
In our design, instead of having different designs for initial and updated templates, we treat them equally.
This may not be the best choice to pursue higher performance.
Empirically, designing different branches for different templates may give better results.
However, we find that the simple method to treat templates as equal works well.
To keep the framework concise, we do not employ extra designs for different templates.
This choice brings an additional advantage in that the model can take different numbers of templates as input without redesigning or retraining, making it flexible in application.
The users can freely select the single-template setup for faster speed or multi-template setup for better performance without retraining.
This characteristic is also due to the following reasons. 
First, attention can receive different numbers of feature vectors as inputs.
The results of the attention layer are calculated by the $softmax$ weighted average using the attention score.
For example, the result of inputting one template is the same as that of inputting multiple copies of this template.
Second, we use the concatenating method to combine the multi-template features.
In the concatenation method, different feature vectors are independent, thereby ensuring the attention layers have seen independent initial-template feature vectors and updated-template feature vectors in training.
Thus, the model learned how to deal with these two types of feature vectors.
In testing, even if the number of templates is different, each template feature vector is no more than one of these two types.
Third, under different training and testing numbers of templates, the proportions of the initial-template feature vectors are different, which creates some training-testing inconsistency.
However, in the experiments, we find that the impact of such inconsistency can be negligible.
One underlying reason is that the updated templates in training are generated by random jitter, which introduces much diversity in training template vectors.
In Section~\ref{sec-exp}, we provide the results of the TransT-M training and testing with different template numbers.

{\noindent \textbf{IoU Prediction Head.}}
The tracker needs an indicator to control the update of the template.
However, the confidence score is not a good indicator because it represents the possibility of the target existing at each feature vector but not whether this feature vector's predicted bounding box is suitable for being a template.
Thus, we design an IoU prediction head to determine whether the prediction is suitable as a template.
The IoU prediction head is simple. We concatenate the regression vectors and the fusion vectors in the channel dimension and use a three-layer perceptron to predict the IoU on each vector. Then, we compare the predicted IoU with the update threshold to control the update of the templates.

\subsection{Training Loss}
{\noindent \textbf{Classification and regression prediction loss.}} 
The prediction head receives $H_x \times W_x$ feature vectors and outputs $H_x \times W_x$ binary
classification and regression results.
We select the prediction of feature vectors corresponding to pixels in the ground-truth bounding box as positive samples, and the rest are negative samples.
All samples contribute to the classification loss, whereas only positive samples contribute to the regression loss.
To reduce the imbalance between positive and negative samples, we downweighted the loss produced by negative samples by a factor of 16.
We employ the standard binary cross-entropy loss for classification, which is defined as
\begin{equation}
\begin{split}
\label{equation:BCE_loss}
\mathcal{L}_{cls} = -\sum_j[y_j{\rm log}(p_j)+(1-y_j){\rm log}(1-p_j)], 
\end{split}
\end{equation}
where $y_j$ denotes the ground-truth label of the $j$-th sample, $y_j=1$ denotes the foreground, and $p_j$ denotes the probability of belonging to
the foreground predicted by the learned model.
For regression, we employ a linear combination of $\ell_1$-norm loss $\mathcal{L}_{1}(.,.)$
and the generalized IoU loss $\mathcal{L}_{GIoU}(.,.)$~\cite{GIoU}.
The regression loss can be formulated as
\begin{equation}
\begin{split}
\label{equation:bbox_loss}
\mathcal{L}_{reg} = \sum_j\mathbbm{1}_{\{y_j=1\}}[\lambda_{G}\mathcal{L}_{GIoU}(b_j,\hat{b})+\lambda_1\mathcal{L}_1(b_j,\hat{b})]
\end{split}, 
\end{equation}
where $y_j=1$ denotes the positive sample, $b_j$ denotes the $j$-th predicted bounding box, and 
$\hat{b}$ denotes the normalized ground-truth bounding box.
$\lambda_{G} =2 $ and $\lambda_{1} = 5$ are the regularization parameters in our experiments. 

{\noindent \textbf{IoU prediction loss.}}
The IoU prediction head outputs $H_x \times W_x$ IoUs. We calculate each positive sample's IoU with the corresponding ground-truth bounding box as the IoU ground truth.
We use the MSE loss for the IoU prediction head.
Only the positive samples contribute to this loss because we only need the IoU prediction to judge the quality of the bounding box rather than to judge the positive and negative samples.

{\noindent \textbf{Segmentation prediction loss.}}
For the segmentation branch, we use the dice loss~\cite{VNET} and the focal loss~\cite{focal_loss}.

\begin{equation}
\begin{split}
\label{equation:seg_loss}
\mathcal{L}_{seg} = \lambda_{D}\mathcal{L}_{Dice}(m,\hat{m})+\lambda_{f}\mathcal{L}_{focal}(m,\hat{m})
\end{split}, 
\end{equation}
where $m$ is the predicted mask and $\hat{m}$ is the ground truth.
$\lambda_{D} =1 $ and $\lambda_{f} = 1$ are the regularization parameters in our experiments.

\begin{table*}[!t]
\caption{State-of-the-art comparison on LaSOT, TrackingNet, and GOT-10k. The best three results 
are shown in \textbf{\textcolor{red}{red}}, \textbf{\textcolor{blue}{blue}}, and \textbf{\textcolor{cGreen}{green}} fonts.}
\label{tab-sota1}
\vspace{-4mm}
\begin{center}
\resizebox{0.95\linewidth}{!}{
  \setlength{\tabcolsep}{3mm}{  
  \small
\begin{tabular}{l|c|ccc c ccc c ccc}
\toprule
\multirow{2}{*}{Method} & \multirow{2}{*}{Year} &\multicolumn{3}{c}{LaSOT~\cite{LaSOT}}	
&
&\multicolumn{3}{c}{TrackingNet~\cite{trackingnet}}
&
&\multicolumn{3}{c}{GOT-10k~\cite{GOT10K}}\\
    \cline{3-5}
    \cline{7-9}
    \cline{11-13}
& &AUC	&P$_{Norm}$	&P	& &AUC	&P$_{Norm}$	&P &	&AO	&SR$_{0.5}$	&SR$_{0.75}$\\
\midrule[0.5pt]
TransT-M    &ours	&\textcolor{cGreen}{\textbf{65.4}}	&\textcolor{cGreen}{\textbf{73.9}}	&\textcolor{cGreen}{\textbf{69.6}}	 &
&\textcolor{red}{\textbf{82.5}}	&\textcolor{red}{\textbf{87.3}}	&\textcolor{blue}{\textbf{80.0}}	&
&\textcolor{red}{\textbf{74.7}}	&\textcolor{red}{\textbf{85.5}}	&\textcolor{red}{\textbf{71.3}}\\
TransT	&ours &64.9	&73.8	&69.0 &
&81.4	&\textcolor{blue}{\textbf{86.7}}	&\textcolor{red}{\textbf{80.3}}	&
&\textcolor{blue}{\textbf{72.3}}	&\textcolor{blue}{\textbf{82.4}}	&\textcolor{blue}{\textbf{68.2}}\\
TransT-GOT	&ours &-	&-	&- &	&-	&-	&-	& &67.1	&76.8	&60.9\\
\midrule[0.1pt]
ToMP50~\cite{ToMP}	&2022 &\textbf{\textcolor{red}{67.6}} &-	&- &	&81.2 	&\textcolor{cGreen}{\textbf{86.4}}	&78.9 &	&-	&-	&-\\
CSWinTT~\cite{CSWinTT} &2022	&\textbf{\textcolor{blue}{66.2}} &\textbf{\textcolor{red}{75.2}}	&\textbf{\textcolor{red}{70.9}} &	&\textcolor{blue}{\textbf{81.9}}	&\textcolor{blue}{\textbf{86.7}}	&\textbf{\textcolor{cGreen}{79.5}}	& &69.4	&78.9	&\textbf{\textcolor{cGreen}{65.4}}\\
UTT~\cite{UTT} &2022	&64.6 &-	&67.2 &	&79.7	&-	&77.0 &	&67.2	&76.3	&60.5\\
RBO~\cite{RBO} &2022	&55.8 &-	&57.0 &	&-	&-	&-	& &64.4	&76.7	&50.9\\
HybTransT~\cite{HybTransT} &2022	&64.5 &\textcolor{cGreen}{\textbf{73.9}}	&- &	&-	&-	&-	 & &-	&-	&-\\
TransInMo~\cite{TransInMo} &2022	&65.3 &\textbf{\textcolor{blue}{74.6}}	&\textbf{\textcolor{blue}{69.9}} &	&\textcolor{cGreen}{\textbf{81.6}}	&-	&- &	&-	&-	&-\\
ARDiMPsuper~\cite{Alpha-Refine} &2021	&65.3	&73.2	&68.0 &	&80.5	&85.6	&78.3 &	&\textbf{\textcolor{cGreen}{70.1}}	&80.0	&64.2\\
TrDiMP~\cite{Wang_2021_Transformer} &2021	&63.9	&-	&61.4 &	&78.4	&83.3	&73.1	&	&68.8	&\textbf{\textcolor{cGreen}{80.5}}	&59.7\\
TrSiam~\cite{Wang_2021_Transformer}  &2021	&62.4	&-	&60.0	& &78.1	&82.9	&72.7	&	&67.3	&78.7	&58.6\\
STMTrack~\cite{fu2021stmtrack}  &2021	&60.6	&69.3	&63.3 &	&80.3 &85.1	&76.7	&		&64.2	&73.7	&57.5\\
SiamBAN-ACM~\cite{han2021SiamBAN-ACM} &2021 	&57.2	&-	&- &	&75.3	&81.0	&71.2	&	&-	&-	&-\\
SiamGAT~\cite{Guo_2021_CVPR}  &2021	&53.9	&63.3	&53.0 &	&-	&-	&-	&	&62.7	&74.3	&48.8\\
DSTrpn~\cite{shen2021DSTrpn}  &2021	&43.4	&51.3	&- &	&64.9	&58.9	&-	&	&-	&-	&-\\
SiamR-CNN~\cite{SiamRCNN}   &2020	&64.8	&72.2	&- &	&81.2	&85.4	&\textbf{\textcolor{blue}{80.0}}	&	&64.9	&72.8	&59.7\\
Ocean~\cite{Ocean}	&2020	&56.0	&65.1 	&56.6 &	&-	&- 	&- &	&61.1	&72.1	&47.3\\
KYS~\cite{KYS}	    &2020	&55.4	&63.3	&- &	&74.0	&80.0	&68.8 &	&63.6	&75.1	&51.5\\
DCFST~\cite{DCFST}	&2020	&-	&-	&- &	&75.2	&80.9	&70.0 &	&63.8	&75.3	&49.8\\
SiamFC++~\cite{SiamFC++}	&2020	&54.4	&62.3	&54.7 &	&75.4	&80.0	&70.5 &	&59.5	&69.5	&47.9\\
PrDiMP~\cite{PrDiMP} &2020	&59.8	&68.8	&60.8 &	&75.8	&81.6	&70.4 &	&63.4	&73.8	&54.3\\
CGACD~\cite{CGACD}	&2020	&51.8	&62.6	&- &	&71.1	&80.0	&69.3 &	&-	&-	&-\\
SiamAttn~\cite{DSA}	&2020	&56.0	&64.8	&- &	&75.2	&81.7	&- &	&-	&-	&-\\
MAML~\cite{MAML}	&2020	&52.3	&-	&- &	&75.7 	&82.2	&72.5 &	&-	&-	&-\\
D3S~\cite{D3S}	    &2020	  &-	&-	&- &	&72.8	&76.8	&66.4 &	&59.7	&67.6	&46.2\\
SiamCAR~\cite{SiamCAR}	&2020	  &50.7	&60.0	&51.0 &	&-	&-	&-	 & &56.9	&67.0 	&41.5\\
SiamBAN~\cite{SiamBAN}	&2020	&51.4	&59.8	&52.1 &	&-	&-	&- &	&-	&-	&-\\
DiMP~\cite{DiMP}	&2019    	&56.9	&65.0	&56.7 &	&74.0	&80.1	&68.7 &	&61.1	&71.7	&49.2\\
SiamPRN++~\cite{SiamRPNplusplus} &2019	&49.6	&56.9	&49.1 &	&73.3	&80.0	&69.4 &	&51.7	&61.6	&32.5\\
ATOM~\cite{ATOM}	  &2019  	&51.5	&57.6	&50.5 &	&70.3	&77.1	&64.8 &	&55.6	&63.4	&40.2\\
ECO~\cite{ECO}	      &2017  	&32.4	&33.8	&30.1 &	&55.4	&61.8	&49.2 &	&31.6	&30.9	&11.1\\
MDNet~\cite{MDNet}	 &2016   	 &39.7	&46.0	&37.3 &	&60.6	&70.5	&56.5 &	&29.9	&30.3	&9.9\\
SiamFC~\cite{SiameseFC}	&2016	&33.6	&42.0	&33.9 &	&57.1	&66.3	&53.3 &	&34.8	&35.3	&9.8\\
 \bottomrule
\end{tabular}}}
\end{center}
\end{table*}

\section{Experiments}
\label{sec-exp}

\subsection{Implementation Details}

{\noindent \textbf{Offline Training. }}
For TransT, we train our model on the training splits of COCO~\cite{COCO}, TrackingNet~\cite{trackingnet}, LaSOT~\cite{LaSOT},
and GOT-10k~\cite{GOT10K} datasets.
For the video datasets (TrackingNet, LaSOT, and GOT-10k), we directly sample the image pairs from one video sequence
to collect training samples.
For COCO detection datasets, we apply some transformations to the original image to generate image pairs.
Common data augmentation (such as translation and brightness jitter) is applied to enlarge the training set.
The sizes of the search region patch and template patch are $256 \times 256$ and $128 \times 128$, respectively.
The backbone parameters are initialized with
ImageNet-pretrained ResNet-50~\cite{ResNet,ImageNet},
and other parameters of our model are initialized with Xavier init~\cite{Xavier}.
We train the model with AdamW~\cite{AdamW}, setting the backbone's learning rate to 1e-5,
other parameters' learning rate to 1e-4,
and weight decay to 1e-4.
%
We train the network on eight NVIDIA Titan RTX GPUs with 14 sample pairs on each GPU (batch size 112) for 500 epochs with 80,000 sample pairs per epoch.
The learning rate decreases by a factor of 10 after 400 epochs.

For TransT-M, we divide the training process into two stages.
First, we train the multi-template TransT baseline.
We use $M=2$, that is, two templates by default.
For the video datasets, we sample $M+1$ frames in a video sequence to form a sample pair. 
To simulate the online tracking process, we select the head frame or the tail frame to generate the search region. 
We use the frame farthest from the search region frame to generate the initial-template patch and use the middle frames to generate the updated-template patches.
For the COCO detection dataset, we apply transformations to generate image pairs. We randomly select one image to generate the search region patch, another image to generate the initial-template patch, and the rest to generate the updated-template patches.
For the search region patch and the updated templates, we jitter and transform the position and shape of the target box when we obtain the patches to simulate the inaccuracy of template updating in tracking.
For the initial template, we do not apply jittering or transformation because the initial target's bounding box given in online tracking is generally accurate.
The other training settings are the same as those in TransT.
For the second stage, we freeze the parameters of the multi-template TransT baseline and train the IoU prediction head.
For the IoU prediction head, we use the same training datasets as the TransT baseline.
We employ AdamW with a learning rate of 1e-3 and weight decay of 1e-4.
We train the IoU prediction head on two NVIDIA Titan RTX GPUs with a batch size of 256 for 90 epochs with 200 iterations per epoch.
The learning rate decreases by a factor of 10 after every 30 epochs.
We do not train the IoU prediction head with the TransT baseline together because we hope that the ground truth of the IoU prediction head is consistent with the final network's bounding box prediction and unaffected by the bad bounding box prediction at the initial stage of training.
For the segmentation branch, we use YouTube-VOS~\cite{xu2018youtube} and some saliency datasets~\cite{shi2015hierarchical,wang2017learning,yang2013saliency} as Alpha-Refine~\cite{Alpha-Refine} to train.
The training strategy is the same as the IoU prediction head, except that the initial learning rate is set to 1e-2.
We do not train the segmentation branch with the TransT baseline together because the segmentation datasets are relatively small, and we do not want it to be flooded by the tracking datasets.

{\noindent \textbf{Online Tracking. }}
In online tracking, the prediction head outputs 1024 boxes with their confidence scores.
Then, the window penalty is adopted for postprocessing these scores. Specifically, the Hanning window with the shape of $32 \times 32$ is applied to scores, weighted by a parameter $w$ (chosen as 0.49 in this work). The final score $score_w$ can be defined as
\begin{equation}
\begin{split}
\label{equation:window_penalty}
score_w = (1 - w) \times score + w \times score_h
\end{split}, 
\end{equation}
where $score$ is the original score of the tracker's output. $score_h$ is the value of the corresponding position on the Hanning window.
Through window penalty, the confidence scores of feature points far from the target in the previous frames are punished.
Finally, we select the box with the highest confidence score as the tracking result.
For TransT, we always use the initial target's bounding box of the first frame as the template and do not update it. For TransT-M, we use two templates by default, and we always keep the initial template and update other templates. The update of the templates is controlled by the predicted IoU. If the predicted IoU is higher than the update threshold, the templates are updated.

\subsection{Evaluation on TrackingNet, LaSOT and GOT-10k}

In this subsection, we compare our methods with state-of-the-art trackers on the large-scale LaSOT~\cite{LaSOT}, TrackingNet~\cite{trackingnet}, and GOT-10k~\cite{GOT10K} datasets. The results are shown in Table~\ref{tab-sota1}.

\begin{figure}
\centering
    \includegraphics[width=1.0\linewidth]{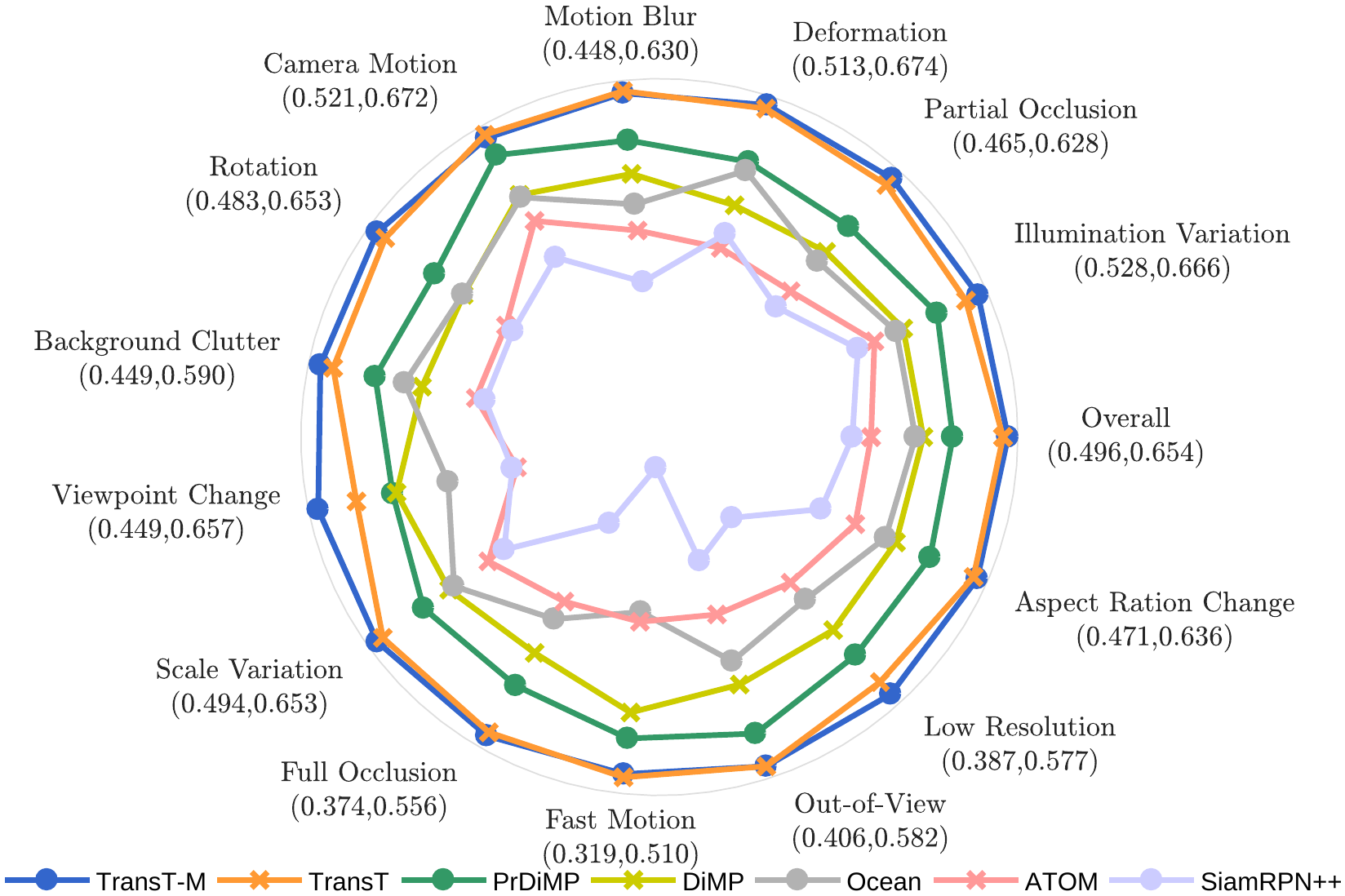}
    \vspace{-5mm}
    \caption{AUC scores of different attributes on the LaSOT dataset.}
    \label{fig:lasotatt}
\end{figure}

{\noindent \textbf{LaSOT. }} LaSOT~\cite{LaSOT} is a recent large-scale dataset, which contains 1,400 challenging videos: 1,120 for training and 280 for testing.
We follow the one-pass evaluation (Success and Precision) to compare different tracking algorithms on the
LaSOT test set.
Then, we report the Success (AUC) and Precision (P and $\rm{P}_{Norms}$) scores in Table~\ref{tab-sota1}.
The proposed TransT-M and TransT achieve competitive performance.
Compared with two transformer-based tackers TrDiMP and TrSiam, our TransT-M method performs 1.5\% and 3.0\% higher, respectively, in AUC.
Figure~\ref{fig:lasotatt} shows an attribute-based evaluation of representative state-of-the-art algorithms, illustrating that our trackers achieve competitive results on many challenging attributes.

{\noindent \textbf{TrackingNet. }} TrackingNet~\cite{trackingnet} is a large-scale tracking dataset that covers diverse object classes and scenes.
Its test set contains 511 sequences of publicly available ground truth.
We submit our tracker's outputs to the official online evaluation server and report the Success (AUC) and Precision (P and P$_{Norm}$) results in Table~\ref{tab-sota1}.
Our TransT obtains 81.4\%, 86.7\%, and 80.3\% in terms of AUC, $\rm{P}_{Norms}$, and P, respectively. TransT-M further improves the performance to 82.5\%, 87.3\%, and 80.0\% in AUC, $\rm{P}_{Norms}$, and P, respectively.
TransT-M achieves the best performance.
Compared with the second-best tracker CSWinTT~\cite{CSWinTT}, TransT-M performs 0.6\% higher in AUC.
Compared with two transformer-based trackers TrDiMP and TrSiam, TransT-M performs 4.1\% and 4.4\% better in AUC.

{\noindent \textbf{GOT-10k. }} The GOT-10k~\cite{GOT10K} dataset contains 10k sequences for training and 180 for testing.
We follow the defined protocol presented in~\cite{GOT10K} and submit the tracking outputs to the official evaluation server.
Then, we report the obtained results (i.e., AO and SR$_{\rm{T}}$) in Table~\ref{tab-sota1}. TransT-GOT denotes training with only the GOT-10k training set.
Our methods achieve the best performance. TransT-M and TransT obtain 74.7\% and 72.3\% AO, respectively, which is much better than previous methods. 
Compared with the third-best tracker ARDiMPsuper~\cite{Alpha-Refine}, TransT-M and TransT perform 4.6\% and 2.2\% better in AO.

\begin{table*}[!ht]
\caption{Ablation study on TrackingNet, LaSOT, and GOT-10k. The best results 
are shown in the \textbf{\textcolor{red}{red}} font.}
\vspace{-6mm}
\label{tab-ab}
\begin{center}
\resizebox{0.95\linewidth}{!}{
  \setlength{\tabcolsep}{4.2mm}{  
  \small
\begin{tabular}{l|ccc c ccc c ccc}
\toprule
\multirow{2}{*}{Method}  &\multicolumn{3}{c}{LaSOT~\cite{LaSOT}}	&
&\multicolumn{3}{c}{TrackingNet~\cite{trackingnet}} &	&\multicolumn{3}{c}{GOT-10k~\cite{GOT10K}}\\
\cline{2-4}
\cline{6-8}
\cline{10-12}
&AUC	&P$_{Norm}$	&P	& &AUC	&P$_{Norm}$	&P &	&AO	&SR$_{0.5}$	&SR$_{0.75}$\\
\midrule[0.5pt]
TransT &\textcolor{red}{\textbf{64.9}}	&\textcolor{red}{\textbf{73.8}}	&\textcolor{red}{\textbf{69.0}}	&
&\textcolor{red}{\textbf{81.4}}	&\textcolor{red}{\textbf{86.7}}	&\textcolor{red}{\textbf{80.3}}	&
&\textcolor{red}{\textbf{72.3}}	&\textcolor{red}{\textbf{82.4}}	&\textcolor{red}{\textbf{68.2}}\\
TransT-np       &62.9	&71.5	&66.9 &	&81.1	&86.4	&80.0	& &71.5	&81.5	&67.5\\
TransT-opp      &63.5	&72.9	&67.4 &	&77.8	&85.5	&80.9	& &69.9	&80.1	&65.0\\
TransT(ori)     &62.3	&71.1	&66.2 &	&81.3	&86.1 	&78.9	& &70.3	&80.2	&65.8\\
TransT(ori)-np  &60.9	&69.4	&64.8 &	&80.9	&85.6	&78.4	& &68.6	&78.2	&65.1\\
TransT(ori)-opp &56.9	&64.6	&58.4 &	&60.6	&67.7	&64.0	& &61.9	&70.4	&55.1\\
\bottomrule
\end{tabular}
}}
\end{center}
\end{table*}

\begin{table*}[!ht]
\caption{Comparison with correlation. The best results 
are shown in the \textbf{\textcolor{red}{red}} font.}
\vspace{-5mm}
\label{tab-correlation}
\begin{center}
\resizebox{0.95\linewidth}{!}{
  \setlength{\tabcolsep}{2.8mm}{  
  \small
\begin{tabular}{l|ccc|ccc c ccc c ccc}
\toprule
\multirow{2}{*}{Method}  &\multirow{2}{*}{ECA} 
&\multirow{2}{*}{CFA} 
&\multirow{2}{*}{Correlation} 
&\multicolumn{3}{c}{LaSOT~\cite{LaSOT}}	&
&\multicolumn{3}{c}{TrackingNet~\cite{trackingnet}} &	&\multicolumn{3}{c}{GOT-10k~\cite{GOT10K}}\\
\cline{5-7}
\cline{9-11}
\cline{13-15}
& & & &AUC	&P$_{Norm}$	&P	& &AUC	&P$_{Norm}$	&P &	&AO	&SR$_{0.5}$	&SR$_{0.75}$\\
\midrule[0.5pt]
TransT &$\surd$ &$\surd$ & &\textcolor{red}{\textbf{64.9}}	&\textcolor{red}{\textbf{73.8}}	&\textcolor{red}{\textbf{69.0}}	&
&\textcolor{red}{\textbf{81.4}}	&\textcolor{red}{\textbf{86.7}}	&\textcolor{red}{\textbf{80.3}}	&
&\textcolor{red}{\textbf{72.3}}	&\textcolor{red}{\textbf{82.4}}	&\textcolor{red}{\textbf{68.2}}\\
TransT   & &$\surd$ &  &62.9	&71.9	&66.2 &	&81.1	&86.2	&79.1 &	&70.6	&81.2	&65.7\\
TransT  &$\surd$ & &$\surd$   &57.7	&65.4	&59.5 &	&77.5	&82.2	&74.0 &	&62.8	&72.2	&54.8\\
TransT  & & &$\surd$   &47.7	&48.6	&41.7 &	&68.8	&71.4	&60.9 &	&50.9	&58.0	&33.3\\
TransT-np  &$\surd$ &$\surd$ &   &62.9	&71.5	&66.9 &	&81.1	&86.4	&80.0 &	&71.5	&81.5	&67.5\\
TransT-np   & &$\surd$ & &61.0	&69.6	&64.5 &	&80.0	&85.0	&77.9 &	&68.1	&78.3	&64.0\\
TransT-np  &$\surd$ &  &$\surd$  &57.3	&65.2	&58.8 &	&76.2	&80.8	&72.8 &	&61.4	&70.7	&53.7\\
TransT-np  & & &$\surd$  &35.3	&17.9	&20.1 &	&46.5	&40.3	&27.4 &	&38.2	&36.8	&7.0\\
\bottomrule
\end{tabular}}}
\end{center}
\end{table*}

\begin{figure}
\centering
    \includegraphics[width=1.0\linewidth]{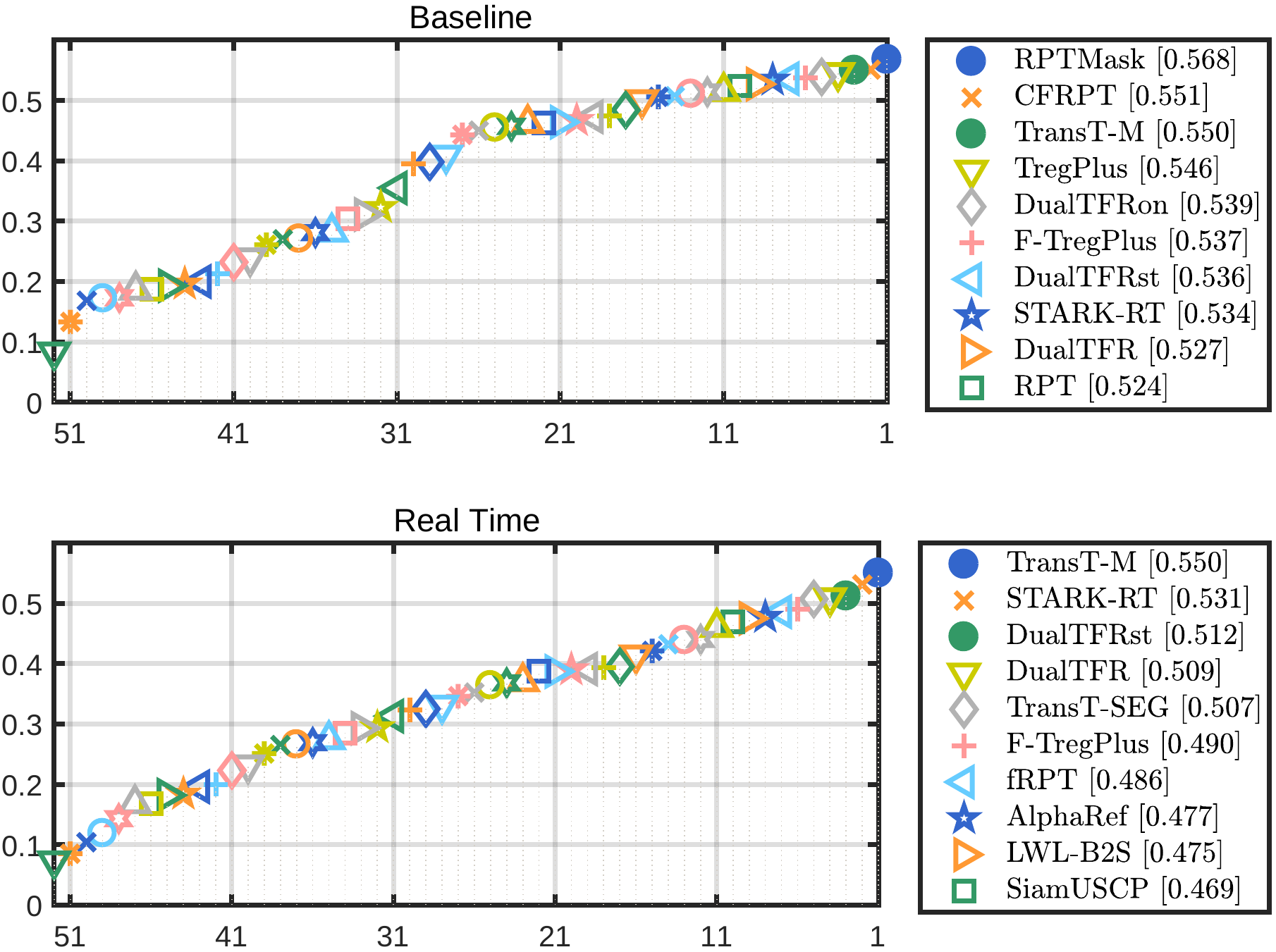}
    \caption{EAO rank plots on VOT2021.}
    \label{fig:votrank}
\end{figure} 

\subsection{Evaluation on VOT}
\label{EvaluationVOT}
We evaluate our tracker on Visual Object Tracking Challenge 2021
(VOT2021)~\cite{Kristan_2021_ICCV} using baseline and real-time experiments.
VOT2021 consists of 60 challenging videos with mask annotation.
VOT2021 adopts expected average overlap (EAO) as the main metric, which simultaneously considers the
trackers’ accuracy and robustness. 
The tracker can submit the mask predictions or bounding box predictions, but the EAO calculation uses the mask annotation.
We use the official evaluation tool and submit the mask prediction.
Figure~\ref{fig:votrank} shows the EAO plots of the proposed methods and other trackers from the official VOT challenge. TransT-SEG is the TransT baseline with our segmentation branch. The results show that our tracker TransT-SEG achieves competitive results in the real-time track.
Furthermore, TransT-M achieves the best result in the real-time track and the third-best result in the baseline track. Thus, it becomes the winner of VOT-RT2021. 
In the real-time track, TransT-M performs 1.9\% better than the second-best tracker STARK~\cite{Stark}, which is also a transformer-based tracker.
The sequence of VOT is difficult, and it has many appearance changes and similar target interference. 
Thus, TransT-M performs much better than TransT-SEG on VOT, due to its strong ability to handle appearance variations.

\subsection{Ablation Study and Analysis}
\label{subsec-as}

\begin{table*}[!ht]
\caption{Comparison with different $N$ of TransT. The best results 
are shown in the \textbf{\textcolor{red}{red}} font.}
\vspace{-6mm}
\label{tab-N}
\begin{center}
\resizebox{0.95\linewidth}{!}{
  \setlength{\tabcolsep}{2.4mm}{  
  \small
\begin{tabular}{l|c|ccc|ccc c ccc c ccc}
\toprule
\multirow{2}{*}{Method}  &\multirow{2}{*}{$N$} 
&Param
&Model Time
&Speed
&\multicolumn{3}{c}{LaSOT~\cite{LaSOT}}	&
&\multicolumn{3}{c}{TrackingNet~\cite{trackingnet}} &	&\multicolumn{3}{c}{GOT-10k~\cite{GOT10K}}\\
\cline{6-8}
\cline{10-12}
\cline{14-16}
& &(M) &(ms)  &($fps$) &AUC	&P$_{Norm}$	&P &	&AUC	&P$_{Norm}$	&P &	&AO	&SR$_{0.5}$	&SR$_{0.75}$\\
\midrule[0.5pt]
TransT   &2 &16.7  &12 &71	&64.2	&73.5	&68.2 &	&80.9	&86.4	&79.2 &	&69.9	&80.1 &65.9\\
TransT   &4 &23.0  &17 &54	&\textcolor{red}{\textbf{64.9}}	&\textcolor{red}{\textbf{73.8}}	&\textcolor{red}{\textbf{69.0}}	& &81.4	&\textcolor{red}{\textbf{86.7}}	&\textcolor{red}{\textbf{80.3}} &	&\textcolor{red}{\textbf{72.3}}	&\textcolor{red}{\textbf{82.4}} &\textcolor{red}{\textbf{68.2}}\\
TransT   &6 &29.3  &22 &42	&64.4	&73.2	&68.5 &	&\textcolor{red}{\textbf{81.6}}	&86.2	&79.6 &	&70.9	&80.4 &66.5\\

\bottomrule
\end{tabular}}}
\end{center}
\end{table*}

\begin{table*}[!ht]
\caption{Comparison with different backbones of TransT. The best results 
are shown in the \textbf{\textcolor{red}{red}} font.}
\vspace{-6mm}
\label{tab-backbone}
\begin{center}
\resizebox{0.95\linewidth}{!}{
  \setlength{\tabcolsep}{2.0mm}{  
  \small
\begin{tabular}{l|c|ccc|ccc c ccc c ccc}
\toprule
\multirow{2}{*}{Method}  &\multirow{2}{*}{Backbone} 
&Param
&Model Time
&Speed
&\multicolumn{3}{c}{LaSOT~\cite{LaSOT}}	&
&\multicolumn{3}{c}{TrackingNet~\cite{trackingnet}}&	&\multicolumn{3}{c}{GOT-10k~\cite{GOT10K}}\\
\cline{6-8}
\cline{10-12}
\cline{14-16}
& &(M) &(ms) &($fps$) &AUC	&P$_{Norm}$	&P	&&AUC	&P$_{Norm}$	&P	&&AO	&SR$_{0.5}$	&SR$_{0.75}$\\
\midrule[0.5pt]
TransT   &ResNet18 &17.1 &15 &61 &63.9	&72.9	&67.7&	&79.8	&85.2	&78.0&	&71.2	&81.8	&65.7\\
TransT   &ResNet50 &23.0 &17 &54	&\textcolor{red}{\textbf{64.9}}	&\textcolor{red}{\textbf{73.8}}	&\textcolor{red}{\textbf{69.0}}	&&81.4	&\textcolor{red}{\textbf{86.7}}	&\textcolor{red}{\textbf{80.3}}&	&\textcolor{red}{\textbf{72.3}}	&\textcolor{red}{\textbf{82.4}} &\textcolor{red}{\textbf{68.2}}\\
TransT   &ResNet101 &42.0 &21 &43 &64.2	&72.8	&68.2&	&\textcolor{red}{\textbf{81.5}}	&\textcolor{red}{\textbf{86.7}}	&80.1&	&72.1	&81.8	&67.8\\

\bottomrule
\end{tabular}}}
\end{center}
\end{table*}

{\noindent \textbf{Postprocessing. }}
In prior works, such as SiamRPN~\cite{SiameseRPN}, SiamRPN++~\cite{SiamRPNplusplus} and Ocean~\cite{Ocean}, the final tracking results are selected by postprocessing schemes, including cosine window penalty, scale change penalty, and bounding box smoothing.
However, these postprocessing schemes are parameter-sensitive because three hyperparameters need to be adjusted carefully for different test sets.
To avoid this problem, in this work, we merely adopt the window penalty to conduct postprocessing
using the default parameter for all test sets.
To show the effect of postprocessing, we compare the TransT variants with and without postprocessing
step in Table~\ref{tab-ab}. TransT denotes our tracker, and TransT-np is our tracker without postprocessing.
First, from Table~\ref{tab-ab}, we can conclude that our TransT
without postprocessing still achieves state-of-the-art performance, which is attributed to the transformer-like
fusion method.
Second, the postprocessing step further improves the tracking accuracy, producing the best record among almost all
metrics on these benchmarks.

\begin{table*}[t]
\caption{Comparison with different $N$ and different $M$ of TransT-M. The best results 
are shown in the \textbf{\textcolor{red}{red}} font.}
\vspace{-6mm}
\label{tab-N&M}
\begin{center}
\resizebox{0.95\linewidth}{!}{
  \setlength{\tabcolsep}{2.0mm}{  
  \small
\begin{tabular}{l|cc|ccc|ccc c ccc c ccc}
\toprule
\multirow{2}{*}{Method}  &\multirow{2}{*}{$N$} 
&\multirow{2}{*}{$M$}
&Param
&Model Time
&Speed
&\multicolumn{3}{c}{LaSOT~\cite{LaSOT}}	&
&\multicolumn{3}{c}{TrackingNet~\cite{trackingnet}}&	&\multicolumn{3}{c}{GOT-10k~\cite{GOT10K}}\\
\cline{7-9}
\cline{11-13}
\cline{15-17}
& & &(M) &(ms) &($fps$) &AUC	&P$_{Norm}$	&P&	&AUC	&P$_{Norm}$	&P&	&AO	&SR$_{0.5}$	&SR$_{0.75}$\\
\midrule[0.5pt]
TransT-M   &2 &1 &16.8 &12 &71  &64.6	&73.7	&68.6&	&81.4	&86.4	&79.4&	&71.3	&81.5	&66.0\\
TransT-M   &2 &2 &16.8 &13 &56 &64.6	&73.7	&68.6&	&81.7	&86.3	&79.0&	&73.5	&84.4	&68.3\\
TransT-M   &2 &3 &16.8 &14 &52  &64.6	&73.7	&68.6&	&81.6	&86.3	&78.8&	&73.1	&84.0	&67.9\\
TransT-M   &4 &1 &23.1 &17 &54 &\textcolor{red}{\textbf{65.4}}	&\textcolor{red}{\textbf{73.9}}	&\textcolor{red}{\textbf{69.6}}	&&81.7	&86.4	&79.8&	&72.7	&82.9	&69.0\\
TransT-M   &4 &2 &23.1 &18 &42 &\textcolor{red}{\textbf{65.4}}	&\textcolor{red}{\textbf{73.9}}	&\textcolor{red}{\textbf{69.6}}&	&\textcolor{red}{\textbf{82.5}}	&\textcolor{red}{\textbf{87.3}}	&80.0&	&\textcolor{red}{\textbf{74.7}}	&\textcolor{red}{\textbf{85.5}}	&\textcolor{red}{\textbf{71.3}}\\
TransT-M   &4 &3 &23.1 &21 &38  &\textcolor{red}{\textbf{65.4}}	&\textcolor{red}{\textbf{73.9}}	&\textcolor{red}{\textbf{69.6}}&	&82.1	&86.9	&79.5&	&74.0	&84.7	&70.5\\
TransT-M   &6 &1 &29.4 &22 &42 &64.2	&72.6	&67.8&	&81.8	&86.8	&\textcolor{red}{\textbf{80.4}}&	&72.8	&83.4	&68.2\\
TransT-M   &6 &2 &29.4 &24 &35 &64.2	&72.6	&67.8&	&82.1	    &87.1	    &80.2	&    &74.3	&85.4	&70.0\\
TransT-M   &6 &3 &29.4 &27 &31 &64.2	&72.6	&67.8&	&82.0	&87.0	&79.6&	&73.6	&84.4	&69.3\\

\bottomrule
\end{tabular}}}
\end{center}
\end{table*}

\begin{table*}[t]
\caption{Comparison with different backbones and different $M$ of TransT-M. The best results 
are shown in the \textbf{\textcolor{red}{red}} font.}
\vspace{-6mm}
\label{tab-backbone&M}
\begin{center}
\resizebox{0.95\linewidth}{!}{
  \setlength{\tabcolsep}{1.7mm}{  
  \small
\begin{tabular}{l|cc|ccc|ccccccccccc}
\toprule
\multirow{2}{*}{Method}  &\multirow{2}{*}{backbone} 
&\multirow{2}{*}{$M$} 
&Param
&Model Time
&Speed
&\multicolumn{3}{c}{LaSOT~\cite{LaSOT}}	&
&\multicolumn{3}{c}{TrackingNet~\cite{trackingnet}}&	&\multicolumn{3}{c}{GOT-10k~\cite{GOT10K}}\\
\cline{7-9}
\cline{11-13}
\cline{15-17}
& & &(M) &(ms) &($fps$) &AUC	&P$_{Norm}$	&P	&&AUC	&P$_{Norm}$	&P&	&AO	&SR$_{0.5}$	&SR$_{0.75}$\\
\midrule
TransT-M   &ResNet18 &1 &17.2 &15 &61 &64.1	&73.6	&67.5&	&80.3	&85.4	&78.2&	&70.9	&81.7	&65.2\\
TransT-M   &ResNet18 &2 &17.2 &16 &50 &64.1	&73.6	&67.5&	&80.6	&85.8	&78.0&	&71.8	&82.9	&65.2\\
TransT-M   &ResNet18 &3 &17.2 &19 &45 &64.1	&73.6	&67.5&	&79.8	&84.9	&76.6&	&71.8	&82.8	&65.2\\
TransT-M   &ResNet50 &1 &23.1 &17 &54 &\textcolor{red}{\textbf{65.4}}	&\textcolor{red}{\textbf{73.9}}	&\textcolor{red}{\textbf{69.6}}&	&81.7	&86.4	&79.8&	&72.7	&82.9	&69.0\\
TransT-M   &ResNet50 &2 &23.1 &18 &42 &\textcolor{red}{\textbf{65.4}}	&\textcolor{red}{\textbf{73.9}}	&\textcolor{red}{\textbf{69.6}}&	&\textcolor{red}{\textbf{82.5}}	&\textcolor{red}{\textbf{87.3}}	&80.0&	&\textcolor{red}{\textbf{74.7}}	&\textcolor{red}{\textbf{85.5}}	&\textcolor{red}{\textbf{71.3}}\\
TransT-M   &ResNet50 &3 &23.1 &21 &38 &\textcolor{red}{\textbf{65.4}}	&\textcolor{red}{\textbf{73.9}}	&\textcolor{red}{\textbf{69.6}}	&&82.1	&86.9	&79.5&	&74.0	&84.7	&70.5\\
TransT-M   &ResNet101 &1 &42.1 &21 &43 &63.1	&72.8	&65.9&	&78.7	&84.3	&76.0&	&71.2	&81.5	&66.3\\
TransT-M   &ResNet101 &2 &42.1 &22 &38  &63.1	&72.8	&65.9&	&81.5	&86.7	&\textcolor{red}{\textbf{80.1}}&	&73.3	&82.5	&67.4\\
TransT-M   &ResNet101 &3 &42.1 &25 &35 &63.1	&72.8	&65.9&	&81.8	&87.0	&79.9&	&72.7	&83.7  &68.1\\

\bottomrule
\end{tabular}}}
\end{center}
\end{table*}

{\noindent \textbf{Comparison with the original transformer. }}
To show the superiority of our feature fusion network, we design a tracker using the original transformer.
As shown in Figure~\ref{fig:TransTandOriTransT} and introduced in Section~\ref{disscussions}, we replace the feature fusion network with the original transformer structure and keep the other components unchanged. We input the template feature to the encoder and the search region feature to the decoder.
The comparison results are shown in Table~\ref{tab-ab}.
TransT(ori) denotes the tracker with the original transformer, and TransT(ori)-np is the TransT(ori) method without postprocessing.
First, the TransT(ori)-np variant achieves an AUC score of 60.9\% on LaSOT, an AUC score of 80.9\% on TrackingNet, and an AO score of 68.6\% on GOT-10k, which is also better than many state-of-the-art correlation-based methods. This indicates that the transformer structure works better than the simple correlation operation in dealing with feature fusion.
Second, by observing TransT \emph{v.s.} TransT(ori) and TransT-np \emph{v.s.} TransT(ori)-np,
We can conclude that the proposed transformer performs better than the original transformer structure by a large margin.
In addition, we also see that the postprocessing works for both the TransT and TransT(ori) methods.

We also evaluate inputting the search region feature to the encoder and the template feature to the decoder, which is denoted as TransT(ori)-opp.
In TransT(ori)-opp, the tracker predicts results on each template feature vector; thus, we cannot divide the positive and negative samples according to the ground-truth box and cannot apply the postprocessing.
We set every template feature vector's prediction as a positive sample.
The results show that TransT(ori)-opp is not as good as TransT(ori) for two reasons.
First, the ground-truth matching independent of the location of the search region leads to all vectors tending to predict the same target, which lacks diversity.
All factors can easily give the same wrong prediction.
In contrast, the correct prediction can be selected from the diverse prediction through confidence score and postprocessing in TransT(ori).
Second, this setting lacks template feature enhancement through cross-attention, which is very important. We exchange the input of the search region feature and the template feature of TransT, which is denoted as TransT-opp. In this setting, we find that the performance degradation is not as much as TransT(ori)-opp because it keeps the enhancement of the template feature through cross-attention.

{\noindent \textbf{Comparison with correlation. }} 
Prior Siamese trackers use the cross-correlation operation to compute the similarity between the template feature and search region feature.
Figure~\ref{fig:correlation} (a) illustrates the cross-correlation layer. The template feature is used as a convolution kernel to
convolve with the search region feature to generate a single-channel response map. 
The response map can measure the similarity between the template and search region.
SiamRPN++ proposed a depthwise cross-correlation layer, as shown in Figure~\ref{fig:correlation} (b), and two feature maps with the same number of channels perform the correlation operation channel by channel to generate multi-channel response maps.
The correlation operation plays a critical role in the tracking field.
However, correlation is a linear calculation without parameters. This single feature fusion process leads to semantic loss, which weakens its feature fusion ability.
Compared with correlation-based methods, first, our attention-based method can establish long-distance feature associations, which effectively aggregate the global information of the template and search region. Second, our method outputs features with rich semantic information, not just a similarity map. 
We conduct experiments to compare CFA with correlation and explore the impact of ECA. To make a fair comparison, for the TransT without CFA we keep the FFN in CFA unchanged, remove only the cross-attention layers, and replace the last CFA module with depthwise correlation. The comparison results are shown in Table~\ref{tab-correlation}.
The comparison results show that after replacing CFA with the correlation layer, the performance significantly decreases.
Without ECA, the performance of the tracker drops. Without ECA and CFA, the performance further drops, and the impact of postprocessing becomes greater.
These results show that without attention modules, the localization ability of the tracker significantly decreases, and it needs to rely more on the prior information in postprocessing.

\begin{table*}[!ht]
\caption{Comparison with different numbers of testing and training templates. The best results 
are shown in the \textbf{\textcolor{red}{red}} font.}
\vspace{-6mm}
\label{tab-M&M}
\begin{center}
\resizebox{0.95\linewidth}{!}{
  \setlength{\tabcolsep}{3.2mm}{  
  \small
\begin{tabular}{l|cc|ccc c ccc c ccc}
\toprule
\multirow{2}{*}{Method}  &\multirow{2}{*}{$M_{test}$} 
&\multirow{2}{*}{$M_{train}$}
&\multicolumn{3}{c}{LaSOT~\cite{LaSOT}}	&
&\multicolumn{3}{c}{TrackingNet~\cite{trackingnet}} &	&\multicolumn{3}{c}{GOT-10k~\cite{GOT10K}}\\
\cline{4-6}
\cline{8-10}
\cline{12-14}
& &  &AUC	&P$_{Norm}$	&P	& &AUC	&P$_{Norm}$	&P &	&AO	&SR$_{0.5}$	&SR$_{0.75}$\\
\midrule[0.5pt]
TransT-M   &1 &1  &64.9	&73.8	&69.0 &	&81.4	&86.7	&\textcolor{red}{\textbf{80.3}} &	&72.3	&82.4	&68.2\\
TransT-M   &2 &1  &64.9	&73.8	&69.0 &	&79.3	&84.3	&73.4	& &73.0	&84.7	&67.8\\
TransT-M   &3 &1 &64.9	&73.8	&69.0  &	&76.9	&81.7	&69.4	& &72.3	&84.6	&65.4\\
TransT-M   &1 &2  &\textcolor{red}{\textbf{65.4}}	&\textcolor{red}{\textbf{73.9}}	&\textcolor{red}{\textbf{69.6}}	 & &81.7	&86.4	&79.8 &	&72.7	&82.9	&69.0\\
TransT-M   &2 &2  &\textcolor{red}{\textbf{65.4}}	&\textcolor{red}{\textbf{73.9}}	&\textcolor{red}{\textbf{69.6}} &	&\textcolor{red}{\textbf{82.5}}	&\textcolor{red}{\textbf{87.3}}	&80.0 &	&\textcolor{red}{\textbf{74.7}}	&\textcolor{red}{\textbf{85.5}}	&\textcolor{red}{\textbf{71.3}}\\
TransT-M   &3 &2  &\textcolor{red}{\textbf{65.4}}	&\textcolor{red}{\textbf{73.9}}	&\textcolor{red}{\textbf{69.6}} &	&82.1	&86.9	&79.5 &	&74.0	&84.7	&70.5\\
TransT-M   &1 &3  &64.2	&73.0	&67.9 &	&82.0	&86.8	&80.2	 & &70.9	&81.2	&65.8\\
TransT-M   &2 &3   &64.2	&73.0	&67.9 &	&82.3	&87.0 &79.8 & 	&74.1	&85.2	&69.3\\
TransT-M   &3 &3  &64.2	&73.0	&67.9 &	&81.6	&86.1	&78.5	 & &73.8	&84.4	&68.4\\
\bottomrule
\end{tabular}}}
\end{center}
\vspace{-2mm}
\end{table*}

\begin{table*}[!ht]
    \caption{Comparison with different template combination methods. The best results 
    are shown in the \textbf{\textcolor{red}{red}} font.}
    \label{tab-avg&concat}
    \vspace{-6mm}
    \begin{center}
\resizebox{0.95\linewidth}{!}{
  \setlength{\tabcolsep}{1.3mm}{  
  \small
    \begin{tabular}{l|ccc|ccc|ccc c ccc c ccc}
    \toprule
    \multirow{2}{*}{Method}  &\multirow{2}{*}{Train} 
    &\multirow{2}{*}{Test}
    &\multirow{2}{*}{$M_{test}$} 
    &Param
    &Model Time
    &Speed
    &\multicolumn{3}{c}{LaSOT}	&
    &\multicolumn{3}{c}{TrackingNet}&	&\multicolumn{3}{c}{GOT-10k}\\
    \cline{8-10}
    \cline{12-14}
    \cline{16-18}
    & & & &(M) &(ms) &($fps$) &AUC	&P$_{Norm}$	&P	&&AUC	&P$_{Norm}$	&P&	&AO	&SR$_{0.5}$	&SR$_{0.75}$\\
\midrule[0.5pt]
    TransT-M   &$Concat$ &$-$ &1 &23.1 &17 &54  &\textcolor{red}{\textbf{65.4}}	&\textcolor{red}{\textbf{73.9}}	&\textcolor{red}{\textbf{69.6}}&	&81.7	&86.4	&79.8	&&72.7	&82.9	&69.0\\
    TransT-M   &$Concat$ &$Concat$ &2 &23.1 &18 &42 &\textcolor{red}{\textbf{65.4}}	&\textcolor{red}{\textbf{73.9}}	&\textcolor{red}{\textbf{69.6}}&	&\textcolor{red}{\textbf{82.5}}	&\textcolor{red}{\textbf{87.3}}	&\textcolor{red}{\textbf{80.0}}&	&\textcolor{red}{\textbf{74.7}}	&\textcolor{red}{\textbf{85.5}}	&\textcolor{red}{\textbf{71.3}}\\
    TransT-M   &$Concat$ &$Avg$ &2 &23.1 &17 &54 		&\textcolor{red}{\textbf{65.4}}	&\textcolor{red}{\textbf{73.9}}	&\textcolor{red}{\textbf{69.6}}&	&82.0	&86.8	&79.6 &&73.1 &83.7	&69.5\\
    TransT-M   &$Avg$ &$-$ &1 &23.1 &17 &54 	&63.7	&73.5	&67.5&	&79.5	&85.3	&77.5  & &69.0	 &79.8	&62.9\\
    TransT-M   &$Avg$ &$Avg$ &2 &23.1 &17 &54 &63.7	&73.5	&67.5	&&79.9	&85.7	&77.6  &&70.5	&82.1 &63.9\\
\bottomrule
    \end{tabular}}}
    \end{center}
    \vspace{-2mm}
\end{table*}

\begin{table}[!ht]
\caption{Ablation study on VOT2021. The best results 
are shown in \textbf{\textcolor{red}{red}}.}
\vspace{-5mm}
\label{tab-vot}
\begin{center}
\resizebox{0.95\linewidth}{!}{
  \setlength{\tabcolsep}{4mm}{  
  \small
\begin{tabular}{l|c|ccc}
\toprule
Method &Seg &EAO &A	&R\\
\midrule[0.5pt]
TransT     &-        &30.0    &47.0	 &76.8\\
TransT-SEG &SEG     &50.7    &\textcolor{red}{\textbf{74.8}}	 &81.7\\
TransT-M   &SEG-na   &53.7    &73.9	 &86.2\\
TransT-M   &SEG      &\textcolor{red}{\textbf{55.0}}	  &74.2	 &\textcolor{red}{\textbf{86.9}}\\
\bottomrule
\end{tabular}}}
\end{center}
\vspace{-4mm}
\end{table}

{\noindent \textbf{Different $N$. }} 
We explore the effect of $N$, i.e., the number of feature fusion layer repeats. The results are shown in Table~\ref{tab-N}. 
\emph{Param} indicates the number of parameters, \emph{Model Time} represents the time of one forward propagation of the model, and \emph{Speed} represents the actual running speed of the algorithm. 
For $N=2,4,6$, the number of parameters
are 16.7, 23.0 and 29.3 M, respectively, and their speeds are 71, 54, and 42 $fps$, respectively, on an NVIDIA Titan RTX GPU.
We found that $N=4$ is better than $N=2$, but $N=6$ is slightly worse than $N=4$ in general.
In training, the final loss of $N=6$ is lower than $N=4$, and the lower performance on test sets is due to overfitting. 
The same phenomenon occurs when we use a larger backbone.
Thus, we choose $N=4$ as the default setting to balance model size, speed, and performance. 
The $N=2$ version also achieves promising results with a higher speed of 71 $fps$.

{\noindent \textbf{Different backbones. }}
The results of employing different backbones are shown in Table~\ref{tab-backbone}.
Param indicates the number of parameters, Model Time represents the time of one forward propagation of the model, and speed represents the actual running speed of the algorithm.
The results show that the tracker with ResNet18 also achieves good performance with fewer parameters and faster speed. However, the performance of the tracker with ResNet101 drops slightly because of overfitting.
For ResNet18, ResNet50, and ResNet101, the number of parameters
are 17.1, 23.0, and 42.0 M, respectively, and their speeds are 61, 54, and 43 $fps$, respectively, on an NVIDIA Titan RTX GPU.
We employ ResNet50 as the default backbone to balance model size, speed, and performance.
The ResNet18 version also obtains good results with a higher speed of $61$ $fp$s.

{\noindent \textbf{Different $M$. }}
We train TranT-M with two templates. However, TransT-M can take different numbers of templates as input without retraining and affecting the performance.
The results of different $M$, that is, the number of templates, are shown in Tables~\ref{tab-N&M} and~\ref{tab-backbone&M}.
Param indicates the number of parameters, model time represents the time of one forward propagation of the model, and speed represents the actual running speed of the algorithm.
For the long-term LaSOT dataset, we do not update the templates.
We also provide the results of TransT-M using different $N$ and backbone networks in these two tables. When we use different $M$, we do not retrain the model.
First, TransT-M with $M=2$ is better than $M=1$ on all datasets with all $N$ and backbone network settings.
This shows that the tracker can benefit from multiple templates, especially when the tracker itself does not perform well, such as TransT-M with ResNet101.
TransT-M records the latest target appearance information in the online tracking process; thus, it is robust to appearance changes.
Second, we observe that the performance of $M=3$ is better than $M=1$ but slightly worse than $M=2$.
The reason is that more updated templates will create more inaccuracies, resulting in performance degradation.
Thus, we choose $M=2$ as the default setting.
Finally, when $M=1$, TransT-M achieves basically the same performance as TransT without retraining, making the application of TransT-M flexible.
The users can freely select the value of $M$ to balance the performance and the speed without retraining.

{\noindent \textbf{Inconsistency of training-testing number of templates.}}
As mentioned above, we train TransT-M with two templates.
A natural question is whether different numbers of training templates influence the performance of different numbers of testing templates.
For example, the three-template setup is worse than the two-template setup during testing, which may be caused by the inconsistency of the training-testing number of templates. If the impact of this inconsistency is small, then we can use the one-template setup for training to reduce the training overhead.
To explore this issue, we conduct experiments with different numbers of training and testing templates.
The results are reported in Table~\ref{tab-M&M}.
Here, we use $M_{test}$ and $M_{train}$ to replace $M$ to avoid confusion.
$M_{test}$ is the number of templates used in testing, and $M_{train}$ is the number of templates used in training.
From the results, first, we see that the single-template (1-template) training setup works well when the testing number is 1, but it cannot benefit from the multi-template testing.
This is because the model trained with the single-template setup never learned the updated templates during training, which makes it cannot adapt to the multi-template testing.
Thus, the training-testing inconsistency has a large influence on single-template training.
Second, for the multi-template training (2-template and 3-template for training), we find that the 2-template setup for training achieves the best performance.
We also observe that the 2-template testing is the best for both the 2-template and 3-template training setups.
These results show that the effect of training-testing inconsistency appears to be small for multi-template setups, and the number of templates itself has a larger influence.
Single-template training cannot adapt to multi-template testing, while 3-template training and testing reduce performance because of more imprecision.
Overall, the 2-template training setup and the 2-template testing setup are good candidates in our experiments.
Under this setting, our TransT-M model can take a different number of templates as input without retraining, and the users can freely choose to use the single-template setup for faster speed or the 2-template setup for better performance.

{\noindent \textbf{Different template combination methods.}}
By default, we use the concatenating method to combine the multi-template features (i.e., we concatenate the features in the spatial dimension before the feature fusion network).
Here, we also attempt to use averaging instead of concatenating to combine the multi-template features to make our method more efficient.
For the averaging method, we directly average the features of multiple templates and send them to the feature fusion network.
The results are shown in Table~\ref{tab-avg&concat} (in this table, all variants adopt the 2-template training setup).
$Concat$ indicates that we concatenate multi-template features, and $Avg$ indicates that we average multi-template features.
For the 1-template testing, the methods of concatenating and averaging are equivalent; thus, we use ``-" to ignore this part.
The first two lines are our default methods with different numbers of testing templates, using the concatenating method for both training and testing.
In the third line, we use the averaging method in testing while keeping the concatenating method in training.
We observe that the averaging method performs worse than the
concatenating method.
In the fourth and fifth lines, we also change the training method to averaging.
We find that using the averaging method in training further reduces performance.
Thus, we choose the concatenating method to pursue better performance.
Compared with the concatenating method, the main disadvantage of the averaging method is that it inevitably integrates unrelated feature vectors of different templates, although it saves some computational overhead.

{\noindent \textbf{Segmentation Head. }}
We conduct an ablation study of the segmentation head on VOT2021, as shown in Table~\ref{tab-vot}.
VOT2021 uses mask annotation for evaluation.
Thus, TransT only achieves 30.2\% EAO by reporting the bounding box.
After adding the segmentation head (denoted as SEG), the TransT-SEG can achieve 49.9\% EAO.
With the multi-template scheme and the segmentation head, TransT-M further obtains 55.2\% EAO.
We remove the additional attention in the segmentation branch (denoted as SEG-na), and then the performance of TransT-M drops to 53.9\%, as shown in the third line in Table~\ref{tab-vot}.

\begin{table}[t]\normalsize
\caption{Comparison with the state-of-the-art on the OTB, NFS, and UAV123 datasets in terms of overall AUC score. The best three results 
are shown in \textbf{\textcolor{red}{red}}, \textbf{\textcolor{blue}{blue}}, and \textbf{\textcolor{cGreen}{green}} fonts.}
    \vspace{-3mm}
  \centering
\resizebox{0.95\linewidth}{!}{
  \setlength{\tabcolsep}{2mm}{
    \small
    \begin{tabular}{l|ccc}
    \toprule
    Method &NFS~\cite{NFS}&OTB~\cite{OTB2015}&UAV123~\cite{UAV}\\
    \midrule[0.5pt]
    TransT-M &\textbf{\textcolor{blue}{66.2}}&68.9&\textbf{\textcolor{red}{70.9}}\\
    TransT &\textbf{\textcolor{cGreen}{65.7}}&69.4&\textbf{\textcolor{blue}{69.1}}\\
    \midrule[0.1pt]
    TrDiMP~\cite{Wang_2021_Transformer} &\textbf{\textcolor{red}{66.5}}&\textbf{\textcolor{blue}{71.9}}&67.5\\
    STMTrack~\cite{fu2021stmtrack} &-&\textbf{\textcolor{cGreen}{71.1}}&64.7\\
    SiamGAT\cite{Guo_2021_CVPR} &-&71.0&64.6\\
    SiamBAN-ACM\cite{han2021SiamBAN-ACM} &-&\textbf{\textcolor{red}{72.0}}&64.8\\
    DSTrpn\cite{shen2021DSTrpn} &-&64.6&-\\
    PrDiMP\cite{PrDiMP} &63.5&69.6&\textbf{\textcolor{cGreen}{68.0}}\\
    DiMP\cite{DiMP} &62.0&68.4&65.3\\
    SiamRPN++\cite{SiamRPNplusplus} &50.2&69.6&61.3\\
    ATOM\cite{ATOM} &58.4&66.9&64.2\\
    HP-siam\cite{dong2019HP-siam} &52.1&60.1&-\\
    HP-bacf\cite{dong2019HP-siam} &50.3&63.4&-\\
    ECO\cite{ECO} &46.6&69.1&53.2\\
    MDNet\cite{MDNet} &42.2&67.8&52.8\\
    \bottomrule
    \end{tabular}
    }
  }

\label{tab-sota-small}
\end{table}

\subsection{Evaluation on Other Datasets}
We evaluate our tracker on some commonly used small-scale datasets,
including NFS~\cite{NFS}, OTB2015~\cite{OTB2015}, and UAV123~\cite{UAV}.
We also collect some state-of-the-art and baseline trackers for comparison. The results are shown in Table~\ref{tab-sota-small}.

{\noindent \textbf{NFS. }}We evaluate the proposed trackers on the 30 $fps$ version of NFS~\cite{NFS} dataset. The NFS dataset contains challenging videos with fast-moving objects.
Our TransT-M and TransT achieve the second- and third-best performance, respectively.

{\noindent \textbf{OTB2015. }}OTB2015~\cite{OTB2015} contains 100 sequences and 11 challenge attributes.
Table~\ref{tab-sota-small} shows that our method does not achieve top performance on this dataset.
We note that OTB2015 is a small-scale dataset and is easily overfitted, and we have not deliberately tuned the hyperparameters for this dataset.

{\noindent \textbf{UAV123. }}UAV123~\cite{UAV} includes $123$ low-altitude aerial videos and adopts success and precision metrics for evaluation.
As shown in Table~\ref{tab-sota-small}, the proposed method performs the best.

\section{Conclusion}
In this work, we propose a novel, simple, and high-performance tracking framework based on the transformer-like feature fusion network.
The proposed network conducts feature fusion solely using the attention mechanism, which includes an ego-context augment module based on
self-attention and a cross-feature augment module based on cross-attention.
The attention mechanism establishes long-distance feature associations, making the tracker adaptively focus on useful information and extract abundant semantic information.
The proposed fusion network can replace the correlation to fuse the template and search region features, thereby facilitating
object localization, bounding box regression, and instance segmentation.
In addition, a segmentation branch, a multi-template scheme and an IoU prediction head are developed to further improve the tracker's performance.
Numerous experimental results on many benchmarks show that the proposed tracker performs significantly better than the state-of-the-art
methods while running at a real-time speed.

\begin{figure}
\centering
    \includegraphics[width=1.0\linewidth]{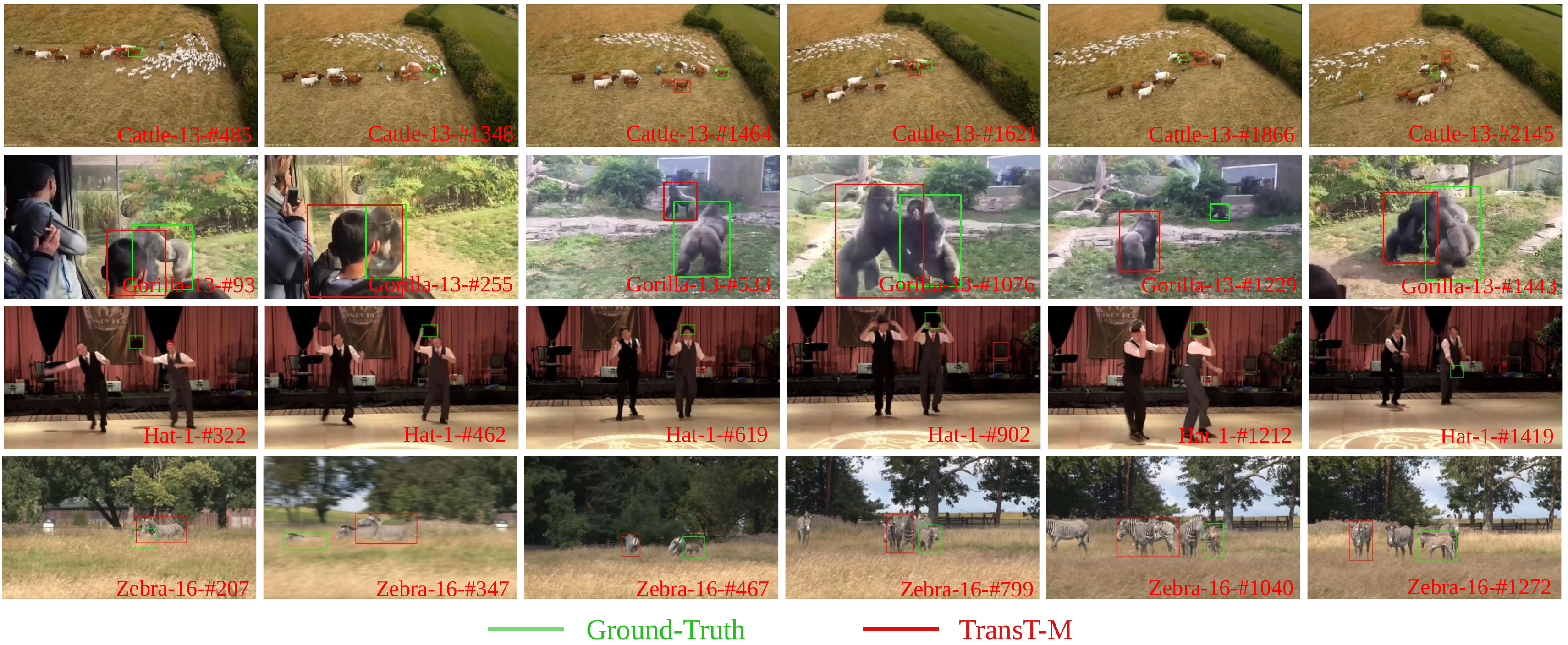}
        \vspace{-8mm}
    \caption{Failure cases. Our method shows difficulty in predicting accurate bounding boxes when the target and distractors are approached. Better viewed in color with zoom-in.}
    \label{fig:failure}
\end{figure}

\textbf{Limitation.}
One limitation of our tracker is that, despite the high performance, it shows difficulty in predicting accurate bounding boxes when the target and distractors are approached. Some failure cases are shown in Figure~\ref{fig:failure}.
Second, our method is developed on a GPU with high computing power and may not run at real-time speed on edge devices, thus limiting its practical application.
Third, while we use the transformer as the feature fusion network, our backbone still employs the CNN-based network. Recently, some works~\cite{sbt,swintrack,ostrack,simtrack,mixformer} demonstrate that employing the transformer as the backbone network can further improve performance.
Moreover, although we design a new architecture for tracking, we do not pretrain the model. The potential of the model may not be fully exploited.

\textbf{Future Work.}
In future work, we are interested in improving our model's efficiency and capacity through the following methods:
1) Employing transformers as the backbone network, 2) developing pretraining methods for tracking, 3) designing lightweight models for edge devices, and 4) incorporating explicit modules to handle distractors.

\clearpage
\bibliographystyle{IEEEtran}
\bibliography{egbib}

\end{document}